\documentclass[preprint,12pt]{elsarticle}
\usepackage{amsmath,amssymb,amsfonts}
\usepackage{array}
\usepackage{amsthm}%
\usepackage{mathrsfs}%
\usepackage{textcomp}
\usepackage{url}
\usepackage{verbatim}
\usepackage{xcolor}%
\usepackage{subfig}
\usepackage{graphicx}%
\usepackage{multirow}%
\usepackage{manyfoot}%
\usepackage{booktabs}%
\usepackage{listings}%
\usepackage[ruled,linesnumbered,lined]{algorithm2e}%
\usepackage[colorlinks=true,linkcolor=blue,citecolor=blue,urlcolor=blue]{hyperref}%
\usepackage{float}
\graphicspath{{./}}
\newcommand{\mstd}[2]{#1{\normalfont\tiny$\pm$#2}}

\begin{document}

\begin{frontmatter}

\title{Decoupled and Distilled: Task-Adaptive LoRA-Teachers with Ensemble Knowledge Transfer for Few-Shot Class-Incremental Learning}

\author[buaa]{Hongwei Zhao}
\ead{zhaohongwei@buaa.edu.cn}
\author[buaa]{Rui Liu}
\ead{lr@buaa.edu.cn}
\author[buaa]{Yansong Liu}
\ead{liuyansong@buaa.edu.cn}
\author[buaa]{Zhiyuan Zou}
\ead{zouzhiyuan@buaa.edu.cn}
\author[bupt]{Yong Chen\corref{cor1}}
\ead{alphawolf.chen@gmail.com}
\cortext[cor1]{Corresponding author.}
\fntext[acceptedmanuscript]{This is the accepted manuscript of an article published in \textit{Neurocomputing}, Article 135200 (2026), \href{https://doi.org/10.1016/j.neucom.2026.135200}{https://doi.org/10.1016/j.neucom.2026.135200}. This manuscript version is made available under the \href{https://creativecommons.org/licenses/by-nc-nd/4.0/}{CC BY-NC-ND 4.0 license}.}

\affiliation[buaa]{organization={School of Computer Science and Engineering, Beihang University},
                  city={Beijing},
                  country={China}}
\affiliation[bupt]{organization={School of Computer Science, Beijing University of Posts and Telecommunications},
                  city={Beijing},
                  country={China}}

\begin{abstract}
Few-Shot Class-Incremental Learning (FSCIL) addresses the fundamental challenge of learning new classes from very limited samples while retaining knowledge of previously learned ones. Although recent parameter-efficient fine-tuning (PEFT) techniques with pre-trained models have shown promise in class-incremental learning, they remain constrained in few-shot settings. In particular, constrained PEFT methods typically impose strict gradient-based constraints to mitigate forgetting; however, maintaining plasticity under such constraints requires abundant training data—a condition unavailable in FSCIL—thereby intensifying the plasticity–stability trade-off. Moreover, multi-expert paradigms that address this plasticity bottleneck introduce substantial inference-time costs through dynamic module selection or runtime module generation, limiting practical deployment.
To address these issues, we propose \textbf{TALON} \textcolor{black}{(\textbf{T}ask-\textbf{A}daptive \textbf{LO}RA-Teachers with E\textbf{N}semble Knowledge Transfer)}, a novel framework that achieves both high accuracy and inference efficiency. TALON introduces task-adaptive LoRA-Teacher modules that are dynamically allocated for each incremental task, enabling effective task-specific representation learning under severe data constraints. To reduce inference overhead, we design an Ensemble Knowledge Transfer (EKT) mechanism that distills knowledge from multiple frozen LoRA-Teachers into a unified LoRA-Student. Furthermore, a semantic-guided distillation strategy adaptively weights teacher contributions based on feature-space similarity, mitigating both catastrophic forgetting and overfitting.
%
\textcolor{black}{Across three class-order runs, TALON achieves \textcolor{black}{comparable or better} mean average accuracy \textcolor{black}{across} all four FSCIL benchmarks, obtaining $86.68\pm1.22\%$ on CUB200, $90.39\pm0.27\%$ on CIFAR100, $78.38\pm0.94\%$ on ImageNet-R, and $96.34\pm0.33\%$ on \textit{mini}ImageNet. TALON maintains clear mean improvements on CIFAR100, ImageNet-R, and \textit{mini}ImageNet, while obtaining performance comparable to SEC-prompt on the more class-order-sensitive CUB200 benchmark. TALON also uses up to $33\times$ fewer deployment parameters and reduces average inference time per task to 26.7~s, corresponding to a 41.70\% reduction relative to ASP (45.8~s).}
Code is available at: \url{https://github.com/hongwei-zhao/NEUCOM-TALON-main}. 
\end{abstract}


\end{frontmatter}

\section{Introduction}
\label{sec:introduction}
In open-world environments, data arrives in a streaming fashion with continually emerging new classes: a scenario known as {Class-Incremental Learning} (CIL). Conventional machine learning models trained sequentially on such data suffer from catastrophic forgetting~\cite{FRENCH1999128,french2020modeling}, where learning new knowledge disrupts previously acquired representations and leads to severe performance degradation.

Real-world applications such as autonomous driving, medical imaging, and robotics~\cite{10350931} often encounter this setting, where novel object categories emerge with only a handful of labeled examples. This gives rise to Few-Shot Class-Incremental Learning (FSCIL)~\cite{9157521}, a particularly challenging paradigm in which models are first trained on base classes with abundant supervision, and then must incrementally adapt to novel classes from severely limited exemplars—typically under an $N$-way $K$-shot protocol~\cite{liu2024few}. This setting exacerbates both catastrophic forgetting of old classes and overfitting to the few new samples, resulting in a pronounced plasticity–stability dilemma~\cite{grossberg2012studies}.

Pre-trained models (PTMs), with their strong generalization abilities~\cite{zhou2024comprehensive}, learned from large-scale datasets~\cite{deng2009imagenet}, provide a strong foundation for incremental learning. However, fully fine-tuning PTMs risks compromising their generalization. Recent advances in CIL address this issue by freezing the PTM backbone and introducing parameter-efficient fine-tuning (PEFT) modules~\cite{xin2024parameter}, such as prompts~\cite{wang2022learning,smith2023coda} or low-rank adaptation (LoRA)~\cite{hu2022lora,gao2023unified,liang2024inflora,wu2025sdlora}. These methods enable task-specific adaptation with minimal trainable parameters, thereby reducing forgetting while preserving generalization.

Despite these advances, existing FSCIL methods still face two fundamental limitations:

\textbf{Challenge I: Insufficient feature learning under data scarcity.} Constrained PEFT-based methods for CIL~\cite{wang2022learning,gao2023unified,liang2024inflora,liu2024few} impose gradient-based constraints (e.g., orthogonal subspaces, shared gradient directions) that require sufficient data to estimate reliably. Under FSCIL's extreme data scarcity, these constraints become unreliable, leading to \emph{underfitting due to insufficient plasticity} or \emph{overfitting due to poor generalization}, depending on the method and dataset (e.g., InfLoRA in Fig.~\ref{fig:motivation}; see the supplementary material for detailed analysis).

\begin{figure}[t]
    \begin{center}
    \includegraphics[width=0.95\columnwidth]{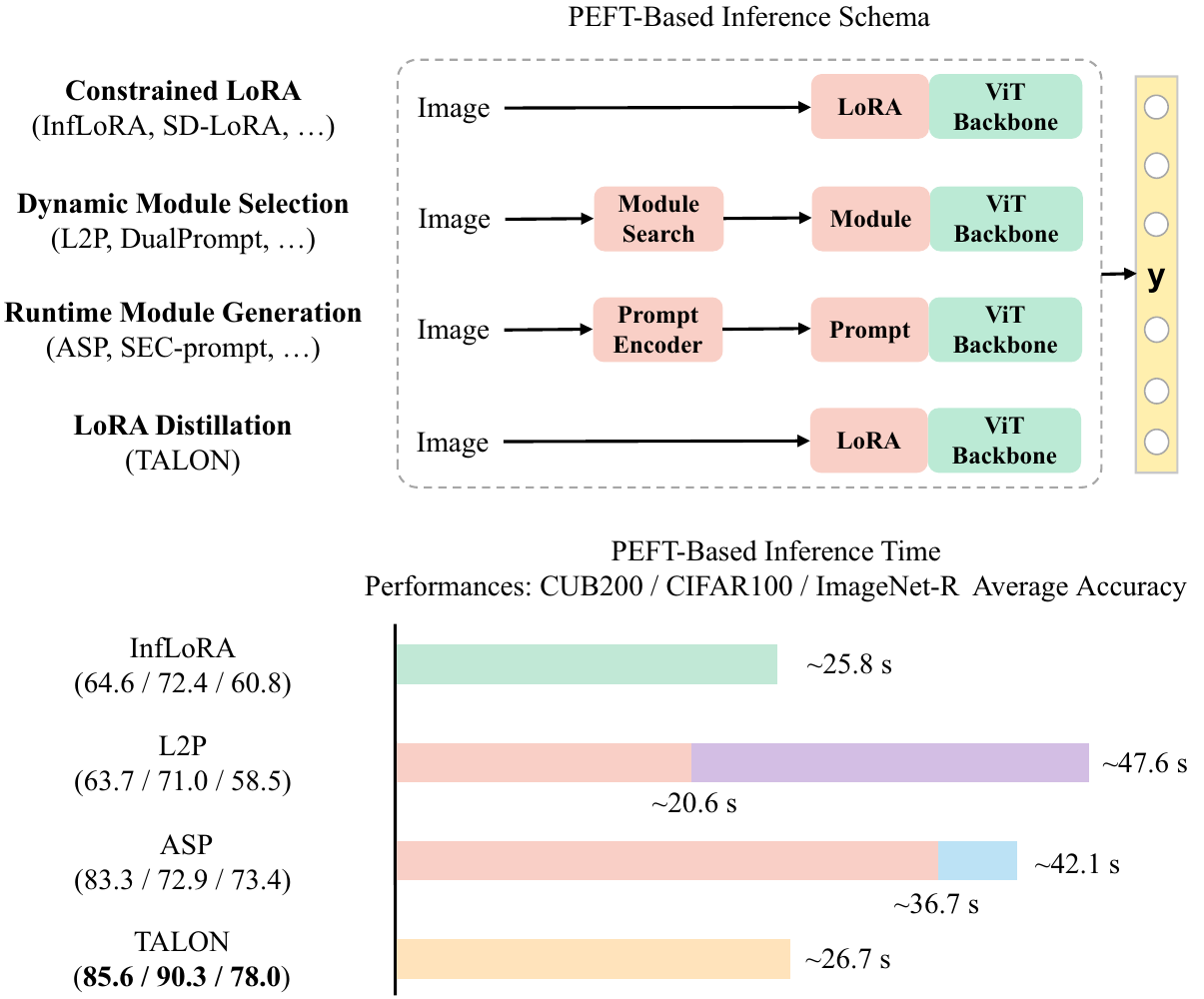}
    \end{center}
    \caption{Performance and inference time comparison among multi-expert-based and LoRA-based methods. TALON consolidates multi-expert knowledge into a single model via distillation, eliminating runtime module selection/generation overhead while achieving superior accuracy (inference time averaged per task on CIFAR100).
    }
    \label{fig:motivation}
\end{figure}

\textbf{Challenge II: Inference inefficiency of multi-expert continual learning paradigms.} To overcome the plasticity bottleneck (Challenge I), a powerful strategy is to use Multi-Expert Learning (e.g., mixture-of-experts or per-sample adapters). However, this introduces substantial inference-time costs:
\begin{itemize}
    \item Dynamic module selection: Query-based methods~\cite{wang2022learning,smith2023coda} require expensive similarity computations across module pools, creating scalability bottlenecks as tasks accumulate.
    \item Runtime module generation: Input-conditional approaches~\cite{liu2024few,liu2025sec} synthesize task-specific modules for each input, incurring per-sample computational overhead.
\end{itemize}

These challenges highlight a central research question: How can we jointly optimize the plasticity–stability trade-off under extreme data scarcity while maintaining inference-time efficiency?

To address this, we propose \textbf{TALON} \textcolor{black}{(\textbf{T}ask-\textbf{A}daptive \textbf{LO}RA-Teachers with E\textbf{N}semble Knowledge Transfer)}, a novel inference-efficient framework for FSCIL that tackles both challenges through a principled multi-teacher distillation strategy.
%
TALON allocates a dedicated LoRA-Teacher module for each incremental task. These teachers are integrated into Transformer layers, enabling efficient representation learning from limited data while maintaining full parameter isolation across tasks—thereby achieving plasticity without sacrificing stability.
To reduce inference overhead, we introduce Ensemble Knowledge Transfer (EKT), which distills collective knowledge from all frozen LoRA-Teachers into a unified LoRA-Student. This design consolidates ensemble-level knowledge into a single model, eliminating dynamic module selection or runtime module generation. Furthermore, we propose a semantic-guided distillation strategy that adaptively weights teacher contributions using feature-space similarity, ensuring coherent and effective knowledge transfer while mitigating both forgetting and overfitting.

Our main contributions are listed as follows:
\begin{itemize}
    \item We identify that the bottleneck of the existing PEFT-FSCIL lies in the `shared-parameter' assumption. We propose a Decoupled Learning paradigm, which dynamically allocates an independent LoRA-Teacher for each incremental task to achieve effective task-specific adaptation under few-shot constraints. Only the current teacher is trained, while previous ones remain frozen, ensuring strong plasticity and stability through full parameter isolation.

    \item We propose Ensemble Knowledge Transfer (EKT), a multi-teacher distillation mechanism that consolidates knowledge from all LoRA-Teachers into a single student, removing the inference-time overhead of dynamic module selection or runtime module generation while retaining ensemble-level performance.

    \item We design a semantic-guided distillation strategy that optimally integrates knowledge from the teacher ensemble based on feature-space similarity, mitigating catastrophic forgetting and overfitting.

    \item Extensive experiments on multiple FSCIL benchmarks show that TALON achieves state-of-the-art performance with superior efficiency. We also demonstrate architectural flexibility through TALON-MLP, a variant exploring alternative teacher integration strategies while maintaining competitive results.
\end{itemize}

\section{Related Work}
\label{sec:related_work}
\subsection{KD-based CIL}
\label{sec:related_work_kd}
Class-Incremental Learning (CIL) aims to continuously acquire knowledge of new classes while retaining previously learned representations~\cite{li2017learning,wang2023comprehensive,qiu2023ism,xi2025pkenet}. Knowledge Distillation (KD)~\cite{KD} is widely used in this context, typically adopting a self-distillation framework where the model trained on prior tasks acts as a teacher, transferring knowledge to the current student model to mitigate forgetting. Following prior work~\cite{10721446}, 3 main KD paradigms have been explored in CIL:

1. KD as regularization~\cite{li2017learning,8953962,asadi2023prototype}: LwF~\cite{li2017learning} pioneered this direction by training the current model to mimic the outputs of the previous model on new data. PRD~\cite{asadi2023prototype} further refined this by continually distilling feature representations using the KL-divergence~\cite{kullback1951information}. Probability dampening and cascaded classifier design have also been combined with KD to balance retention and adaptation~\cite{pomponi2025pdcc}.

2. KD with data replay~\cite{rebuffi2017icarl,Hou_2018_ECCV,9010368}: iCaRL~\cite{rebuffi2017icarl} introduced exemplar replay alongside KD to alleviate forgetting, while PODNet~\cite{douillard2020podnet} extended this with spatially aware feature distillation. K3D jointly optimizes synthetic exemplars and knowledge fusion distillation to preserve class decision boundaries~\cite{xiong2025k3d}. 

3. KD with feature replay~\cite{9578909,10378369,9878745}: For rehearsal-free scenarios, methods such as PASS~\cite{9578909} use class prototypes augmented with Gaussian noise to prevent bias, while Fusion~\cite{9878745} addresses prototype drift by modeling features with Gaussian or variational distributions.

Recent exemplar-free work also combines logit- and feature-level distillation with prototype-based classifier calibration to address forgetting and incremental class bias~\cite{chen2026unidtr}.

\subsection{PEFT-based CIL}
Recent progress in Parameter-Efficient Fine-Tuning (PEFT)~\cite{he2021towards} has shown strong potential for CIL by enabling adaptation with minimal parameters while preserving the generalization of pre-trained models (PTMs). Several representative works include:

1. Prompt-based methods: L2P~\cite{wang2022learning} introduced a dynamic prompt pool to guide PTMs in sequential task learning. DualPrompt~\cite{wang2022dualprompt} separated general prompts (task-invariant) from expert prompts (task-specific). S-Prompts~\cite{wang2022s} further enhanced flexibility by storing domain-specific prompts and selecting them at inference via K-Nearest Neighbors (KNN). CODA~\cite{smith2023coda} proposed decomposed attention-based prompting for rehearsal-free continual learning.

2. LoRA-based methods: LAE~\cite{gao2023unified} employed a learning-accumulation-integration strategy to maintain knowledge across tasks. InfLoRA~\cite{liang2024inflora} reduced task interference by defining task-specific subspaces with injected low-rank parameters. SD-LoRA~\cite{wu2025sdlora} decoupled gradient direction and magnitude, preserving early-task directions while adapting to new ones, albeit at the expense of reduced plasticity.

\subsection{PEFT-based FSCIL}
Few-Shot Class-Incremental Learning (FSCIL) focuses on learning novel classes from extremely limited examples while retaining past knowledge~\cite{9157521,li2023imco}. Early approaches~\cite{kim2023warping,zhang2021few} relied on shallow networks and full backbone fine-tuning, which often caused overfitting and poor generalization~\cite{park2024pre}. More recent work leverages PTMs with PEFT:

1. ASP~\cite{liu2024few} freezes the PTM backbone and uses a prompt encoder to generate task-specific and task-invariant prompts.

2. SEC-prompt~\cite{liu2025sec} designs hierarchical attention queries to generate discriminative and non-discriminative prompts, improving few-shot representation learning.

3. CPE-CLIP~\cite{10350931} integrates learnable prompts into vision–language encoders, combining prompt regularization with CLIP’s cross-modal alignment to mitigate forgetting and enhance transferability.

{\color{black}Recent continual-learning studies have investigated prompting and analytic learning, slow--fast collaborative learning, prototype-enhanced composition, uncertainty-guided expert selection, and domain-specific subspace or prototypical learning for SAR target recognition~\cite{yue2026pal,zhang2026sfcl,li2025pencil,zhang2025chooseexpert,zhao2024aasc,xu2026paepn}. These directions provide complementary perspectives on preserving and organizing knowledge across incremental tasks, while TALON focuses on few-shot image class expansion through task-specific LoRA adaptation and consolidation into a single Student.}

\subsection{Our Approach} 
Building on recent PEFT methods~\cite{liang2024inflora,wu2025sdlora,liu2024few,liu2025sec}, we adopt a PTM backbone with LoRA adaptation~\cite{hu2022lora}. Unlike existing FSCIL approaches that rely on prompt encoders, we propose stage-wise dynamic LoRA allocation, assigning a dedicated LoRA-Teacher to each incremental task for more effective few-shot learning. To consolidate knowledge, we introduce Ensemble Knowledge Transfer (EKT), which distills information from all frozen LoRA-Teachers into a unified student model. A semantic-guided weighting strategy further ensures coherent integration by adapting teacher contributions based on feature-space similarity. This design enables efficient knowledge transfer, robust generalization, and the elimination of inference-time prompt generation. 

\section{Preliminaries}
\label{preliminaries}
\subsection{Problem Formulation}  In Few-Shot Class-Incremental Learning (FSCIL), we consider a sequence of tasks denoted by $\mathcal{D} = \left\{\mathcal{D}_{0}, \cdots, \mathcal{D}_{T}\right\}$, where the $t$-th task $\mathcal{D}_{t}=\left\{\left(\mathbf{x}_{i}, \mathbf{y}_{i}\right)\right\}_{i=1}^{n_t}$ contains $n_t$ instances. Here, $\mathbf{x}_{i} \in \mathcal{X}_t$ represents an input sample from domain $\mathcal{X}_{t}$, and  $\mathbf{y}_{i} \in \mathcal{Y}_t$ is corresponding label from the label space $\mathcal{Y}_t$. Crucially, the label spaces are disjoint across tasks ($\mathcal{Y}_t \cap \mathcal{Y}_{t^\prime}= \emptyset $ for $t \neq t^\prime$). The first task $\mathcal{D}_{0}$ provides ample training data and is referred to as the base task. In contrast, each subsequent incremental task $\mathcal{D}_{t}$ ($t \geq 1$) comprises a limited number of labeled samples, typically formulated as $N$-way $K$-shot classification tasks, where $N$ denotes the number of novel classes introduced and $K$ represents the number of examples per class.

Following the rehearsal-free setting~\cite{wang2022learning,smith2023coda}, the model only has access to the current task’s data during training. Model performance is evaluated on all previously seen classes $\mathcal{Y}_{\le t}=\mathcal{Y}_0 \cup \cdots  \cup \mathcal{Y}_t$ after learning each incremental task. Formally, our objective is to learn a model \textcolor{black}{$f_{\Theta}(\mathbf x)=\mathbf W_{\mathrm{cls}}^\top\phi(\mathbf x)$}  that minimizes the empirical risk over the current training set:
\begin{equation}
    \min_{\Theta} \mathbb{E}_{(\mathbf{x}_i, \mathbf{y}_i) \in \mathcal{D}_{t}} \left[ \text{L} \left(\text{f}_{\Theta}(\mathbf{x}_i), \mathbf{y}_i\right) \right],
    \label{eq:cilrisk}
\end{equation}
\textcolor{black}{where $\phi(\cdot):\mathbb R^D\rightarrow\mathbb R^d$ denotes the feature extractor, $\mathbf W_{\mathrm{cls}}$ denotes a generic linear classifier in the FSCIL problem formulation, and $\text{L}(\cdot,\cdot)$ denotes a loss function measuring the discrepancy between the prediction and the ground-truth label. TALON's training-only shared linear head $\mathbf W_{\mathrm{head}}$ and the deployed prototype bank $\mathcal P_t$ used for classification are introduced separately below.}

\subsection{Low-Rank Adaptation} LoRA~\cite{hu2022lora} was initially introduced for fine-tuning pre-trained models. It achieves fine-tuning of models by adding low-rank matrices $\Delta \mathbf{W}$ to the weight matrices $\mathbf{W} \in \mathbf{R}^{d_{in} \times d_{out}}$ of the pre-trained model. The update rule of the weight matrix $\mathbf{W}$ can be formulated as:
\begin{equation}
   \mathbf{W} + \Delta \mathbf{W} = \mathbf{W} + \mathbf{UV} ,
   \label{e3}
\end{equation}
where $\mathbf{U} \in \mathbf{R}^{d_{in} \times r},\mathbf{V} \in \mathbf{R}^{r \times d_{out}}$ and the rank $r \ll \min(d_{in}, d_{out})$. LoRA significantly reduces the number of trainable parameters, thus improving efficiency and reducing costs in the fine-tuning phase.

\subsection{Knowledge Distillation in Incremental Learning} Knowledge Distillation (KD) is a widely adopted technique that transfers knowledge from a well-trained teacher model to a student model following the teacher-student paradigm. In the context of incremental learning, KD serves as a crucial regularization mechanism to mitigate catastrophic forgetting by preserving knowledge acquired from previous tasks~\cite{10721446}.

In incremental learning scenarios, KD typically employs a self-distillation framework where the teacher and student models share identical architectures. Specifically, the model trained on previous tasks acts as the teacher, transferring its learned representations to the current model (student), thereby maintaining the memory of previously encountered tasks. The knowledge distillation loss $L_{\text{KD}}$ in incremental learning is formulated as:
\begin{equation}
	L_{\text{KD}} = \mathbb{E}_{(\mathbf{x},\mathbf{y}) \in \mathcal{D}_t} \left[ \text{KD}\left(\phi_{t-1}(\mathbf{x}) \parallel \phi_t(\mathbf{x})\right) \right],
\end{equation}
where $\phi_{t-1}$ and $\phi_t$ represent the feature extractors from the old task model and the current task model, respectively, and $\text{KD}(\cdot \parallel \cdot)$ denotes the distillation loss function. This regularization term ensures that the current model $\phi_t$ maintains compatibility with previously learned representations while effectively adapting to new tasks.

\begin{figure*}
  \centering
  \includegraphics[width=1\textwidth]{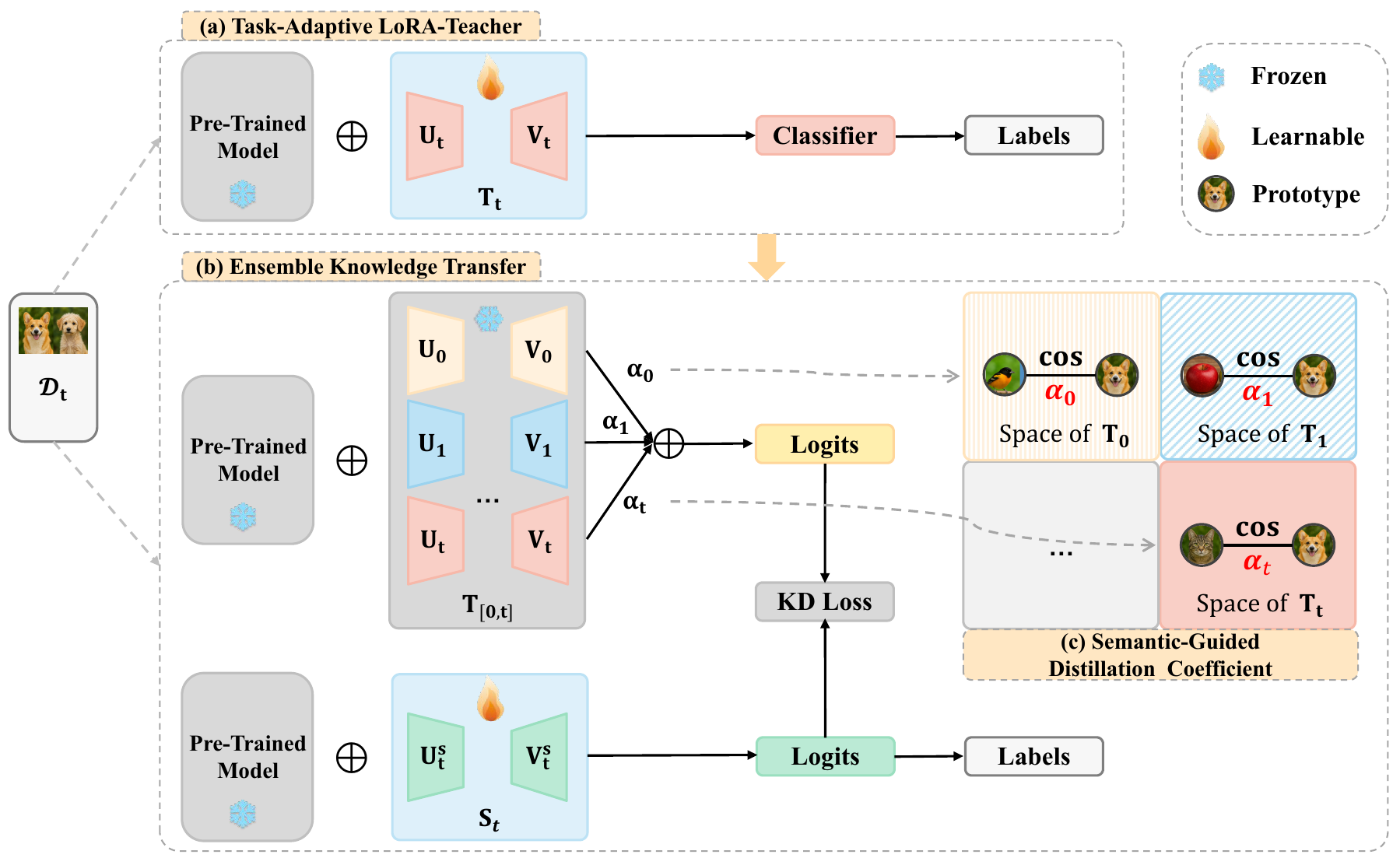}
\caption{Overview of \textcolor{black}{the} TALON framework. \textbf{(a)} Task-Adaptive LoRA-Teacher: For each incremental task $t$, a new LoRA-Teacher $\mathbf{T}_t$ is trained to capture task-specific features while keeping previous \textcolor{black}{LoRA-Teachers} frozen. \textbf{(b)} Ensemble Knowledge Transfer: Multiple LoRA-Teachers are distilled into a unified LoRA-Student $\mathbf{S}_t$ using semantic-guided adaptive coefficients. \textbf{(c)} Semantic-Guided Distillation Coefficients $\alpha_k$: prototypes are extracted from the \textcolor{black}{$k$-th LoRA-Teacher} for task $t$ data using Eq.~(\ref{eq:alpha_prototype}), and the final coefficient $\alpha_k$ is obtained by normalizing raw task-relevance scores in the shared feature space of \textcolor{black}{LoRA-Teacher $\mathbf{T}_k$} according to Eq.~(\ref{eq:distill_coeff}).}
  \label{TALON}
\end{figure*}

\section{The Proposed Method}
\label{sec:methods}
\subsection{Overall Framework}
The overall framework of TALON is illustrated in Fig.~\ref{TALON}. It operates in a two-stage paradigm for each incremental task to effectively learn from limited samples while ensuring high inference efficiency.

\textbf{Stage 1: Task-Adaptive LoRA-Teacher.} Given the limited training samples in few-shot scenarios and strong generalization of pre-trained models, we keep the pre-trained weights $\mathbf{W}$ fixed throughout incremental training. 

To ensure model plasticity and maintain within-task prediction performance at each incremental stage~\cite{wang2023hierarchical}, we adopt a task-isolated LoRA training where a new LoRA-Teacher is dynamically introduced for each task while previous teachers remain frozen.
Specifically, a new LoRA-Teacher is dynamically introduced for each incremental task, while the parameters of previously added LoRA-Teachers remain frozen, shown in Fig.~\ref{TALON}(a). By combining the current task's LoRA-Teacher with pre-trained weights, fine-tuning can be achieved with fewer parameters, enabling efficient capture of task-specific features. 

\textbf{Stage 2: Ensemble Knowledge Transfer (EKT).} To reduce inference overhead, we propose an Ensemble Knowledge Transfer (EKT) framework that distills the collective knowledge from multiple LoRA-Teachers into a unified LoRA-Student model, shown in Fig.~\ref{TALON}(b). This approach enables effective knowledge integration without requiring dynamic module selection or runtime module generation, while maintaining single-model inference efficiency. Unlike existing methods that necessitate complex routing mechanisms ~\cite{wang2022learning} or runtime module generation~\cite{liu2024few,liu2025sec} during inference, our EKT framework consolidates all teacher knowledge into a single student model, thereby ensuring computational efficiency without compromising learning effectiveness.

\textcolor{black}{During Stage~2 EKT, sequential evaluation of the accumulated frozen LoRA-Teachers incurs $O(t)$ Teacher-side forward computation and a linearly growing retained Teacher parameter state. Both costs are training-only: before deployment, all LoRA-Teachers and their Teacher-side prototypes are discarded, leaving only the consolidated LoRA-Student and the accumulated Student prototype bank for inference.}

To alleviate catastrophic forgetting and overfitting during distillation, we introduce the \textbf{Semantic-Guided Distillation Coefficient}, shown in Fig.~\ref{TALON}(c). This strategy adaptively computes coefficients according to the similarity between current and previous category prototypes within the shared feature space. In this way, the student model emphasizes knowledge distilled from teachers most relevant to the current task, thereby improving learning efficiency while reducing conflicting supervision.

\textcolor{black}{For prototype-based inference, new-class prototypes are extracted with the current LoRA-Student while historical prototypes remain fixed; the resulting prototype--feature drift and its decision-level effect are analyzed in Section~\ref{sec:prototype_drift}.}

\subsection{Task-Adaptive LoRA-Teacher} 
To incorporate task-specific information into LoRA modules while mitigating forgetting and minimizing inference overhead, prior studies~\cite{liang2024inflora,wu2025sdlora} have explored gradient-based strategies such as subspace orthogonality and shared gradient directions. However, these approaches rely on sufficient training data to extract meaningful gradient information, which is often infeasible in few-shot scenarios. In contrast, TALON introduces Task-Adaptive LoRA-Teachers that learn directly from limited samples without additional constraints, enabling more effective task-specific adaptation.

We integrate the LoRA-Teacher into the Vision Transformer (ViT) architecture~\cite{dosovitskiy2020image}. In ViT, an input image is first partitioned into fixed-size patches, which are linearly projected and combined with positional embeddings before being processed by the Transformer encoder. The encoder consists of multi-head self-attention (MHA) layers and multilayer perceptron (MLP) blocks.

To adaptively capture task-specific features in incremental learning, we dynamically introduce a LoRA-Teacher module at each incremental stage. This module can be attached as a parallel branch to the MLP block (MLP-Teacher), or to the query ($\mathbf{W}_q$), key ($\mathbf{W}_k$), and value ($\mathbf{W}_v$) projections within the MHA mechanism (QKV-Teacher). When applied jointly to $\mathbf{W}_q$ and $\mathbf{W}_v$, we refer to the configuration as the QV-Teacher.

The modified forward computations for the MLP and MHA layers are formulated as follows:
\begin{equation}
    \mathbf{h}^\prime = \mathbf{e} + \text{MLP}{\left( \mathbf{e} \right)} +  \mathbf{T}_{t}^{MLP}\left({\mathbf{e}}\right),\\
    \label{eq:mlp_teacher}
\end{equation}
\begin{equation}
    \mathbf{h}^\prime =\text{Attn}\left(\mathbf{h}_Q+\mathbf{T}_{t}^{Q}\left({\mathbf{e}}\right),\mathbf{h}_K+\mathbf{T}_{t}^{K}\left({\mathbf{e}}\right),\mathbf{h}_V+\mathbf{T}_{t}^{V}\left({\mathbf{e}}\right)\right),
    \label{eq:qkv_teacher}
\end{equation}
where $\mathbf{e}$ and $\mathbf{h}$ are the input and output of the original module, and $\mathbf{T}_{t}^{Q}$ is the Q-Teacher for task $t$, and \text{Attn} is defined as:
\begin{equation}
    \text{Attn}\left(\mathbf{Q}, \mathbf{K}, \mathbf{V} \right) = \text{softmax}\left(\frac{\mathbf{Q} \mathbf{K}^\top}{\sqrt{d}}\right) \mathbf{V},\\
    \label{eq:attn}
\end{equation}
and the multi-head mechanism is omitted for conciseness. All LoRA-Teachers have the same architecture but with different weights. The LoRA-Teacher consists of a low-rank matrix $\mathbf{U} \in \mathbf{R}^{d_{in} \times r}$ and $\mathbf{V} \in \mathbf{R}^{r \times d_{out}}$, \textcolor{black}{which is formulated as:}
\begin{equation}
    \mathbf{T}_{t}\left({\mathbf{e}}\right) = {\mathbf{e}}\mathbf{U}_t\mathbf{V}_t.
    \label{eq:teacher_output}
\end{equation}

Following LoRA~\cite{hu2022lora}, we initialize $\mathbf{V}$ as a zero matrix and $\mathbf{U}$ using Kaiming initialization~\cite{he2015delving}. Each introduced LoRA-Teacher is initialized in this manner. As illustrated in Fig.~\ref{TALON}(a), for the $t$-th incremental task, a new LoRA-Teacher is learned, while all previously learned LoRA-Teachers remain frozen.

\subsection{Ensemble Knowledge Transfer}
After learning each incremental stage's LoRA-Teacher, we can follow previous strategies to complete inference, such as using routing mechanisms~\cite{wang2023hierarchical} to select the most suitable LoRA-Teacher or obtaining the final inference model through Exponential Moving Average (EMA)~\cite{gao2023unified}. However, both strategies have their respective drawbacks: routing mechanisms increase inference overhead while classification performance is also affected by routing accuracy~\cite{wang2023hierarchical}. The EMA approach requires manually setting hyperparameters to control the EMA decay rate, which is not suitable for FSCIL scenarios.

Model ensemble is a powerful technique to improve model accuracy~\cite{10.1145/3340531.3412704}, yet few works explore its potential in FSCIL. In TALON, we propose an Ensemble Knowledge Transfer (EKT) mechanism that consolidates knowledge from multiple LoRA-Teachers into a unified LoRA-Student through complementary soft logits distillation and hard label distillation strategies.

\textbf{Soft Logits Distillation.} As described in Section~\ref{sec:related_work_kd}, three main KD paradigms exist in CIL: KD as regularization, KD with data replay, and KD with feature replay. Given FSCIL's limited sample constraints, data replay and feature replay become impractical due to insufficient historical data availability. Soft logits convey the subtle difference between two samples and, therefore, can help the student model generalize better than directly learning from hard labels. In this approach, we distill the logits of the student model by minimizing the Kullback-Leibler (KL) divergence~\cite{kullback1951information} between the student and teacher logits distributions. 

Specifically, for a given input $\mathbf{x}$, the soft logits distillation loss is formulated as:
\begin{equation}
    L_{\text{SD}} = \mathbb{E}_{(\mathbf{x},\mathbf{y}) \in \mathcal{D}_t} \left[ \text{L}_\text{KL}\left(\sum_{i=0}^{t}\alpha_i T_i \parallel S_t \right) \right],
    \label{eq:distill_loss}
\end{equation}
where $\text{L}_\text{KL}$ denotes the Kullback-Leibler divergence loss, $\alpha_i$ is the semantic-guided distillation coefficient described in Section~\ref{sec:semantic_coff}, and $T_i$, $S_t$ denote the logits of the $i$-th LoRA-Teacher and $t$-stage LoRA-Student, respectively:
\begin{equation}
    T_i = \mathbf{W}_\text{head}^\top \phi\left(\mathbf{x};\text{T}_i\right), \quad S_t = \mathbf{W}_\text{head}^\top \phi\left(\mathbf{x};\text{S}_t\right),
    \label{eq:teacher_student_features}   
\end{equation}
\textcolor{black}{where $\phi(\cdot;\mathrm{T}_i)$ and $\phi(\cdot;\mathrm{S}_t)$ denote the feature extractors of the $i$-th LoRA-Teacher and the stage-$t$ LoRA-Student, respectively. At session $t$, the shared linear head $\mathbf W_{\mathrm{head}}\in\mathbb R^{d\times|\mathcal Y_{\leq t}|}$ contains one column for each observed class. It maps all Teacher and Student features into the same logit coordinates during EKT. The weighted combination $\sum_{i=0}^{t}\alpha_iT_i$ is computed in logit space before applying softmax. During EKT, $\mathbf W_{\mathrm{head}}$ and all accumulated Teachers are frozen; only the low-rank parameters of $\mathbf S_t$ are optimized. The shared linear head is used only during training and is not used for final inference. At session $t$, the LoRA-Student is initialized from $\mathbf S_{t-1}$, which carries forward the representation consolidated over previous sessions. The relevance-weighted Teacher logits over $\mathcal Y_{\leq t}$ then regularize the low-rank Student update on $\mathcal D_t$, while the hard-label term incorporates the current classes.}


\textbf{Hard Label Distillation.} To ensure the student model maintains strong discriminative capability on the primary classification task, we employ hard label distillation as complementary supervision. This strategy provides explicit ground-truth guidance that balances the implicit knowledge transfer from soft features, preventing the student from over-relying on teacher representations at the expense of task-specific performance. The hard label distillation loss is formulated as:
\begin{equation}
    L_{\text{HD}} = \mathbb{E}_{(\mathbf{x},\mathbf{y}) \in \mathcal{D}_t} \left[ \text{L}_{\text{CE}}\left( \mathbf{W}_\text{head}^\top \phi(\mathbf{x};\text{S}_t), \mathbf{y} \right) \right],
    \label{eq:hard_label_loss}
\end{equation}
where $\text{L}_\text{CE}(\cdot,\cdot)$ denotes the cross-entropy loss between the student's predictions and ground-truth labels.

\textbf{Unified Knowledge Transfer.} The total EKT loss combines both distillation strategies with adaptive weighting:
\begin{equation}
    L_{\text{EKT}} = L_{\text{SD}} + L_{\text{HD}}.
    \label{eq:total_distill_loss}
\end{equation}

This unified approach allows the student model to harness nuanced feature representations from the teacher ensemble while preserving prior knowledge and adapting to new tasks in FSCIL.

\subsection{Semantic-Guided Distillation Coefficient} 
\label{sec:semantic_coff}
Since different LoRA-Teachers exhibit varying relevance to the current task, their contributions to the student model are not uniformly beneficial. In fact, an ineffective teacher may even hinder the student’s learning. To selectively transfer useful knowledge, inspired by EASE~\cite{zhou2024expandable}, we introduce a semantic-guided distillation coefficient that adaptively weights each teacher according to its task relevance. This mechanism enables the student to emphasize the most pertinent knowledge while suppressing less informative or conflicting signals. Concretely, for each teacher $\text{T}_k$, we compute prototypes of its own classes and those of the current task within $\text{T}_k$'s feature space, and measure their raw semantic relevance score:
\begin{equation}
    r_k = \sum_{i=1}^{|\mathcal{Y}_k|}\sum_{j=1}^{|\mathcal{Y}_t|} \left( \frac{\mathbf{P}_{k,k}[i]}{\|\mathbf{P}_{k,k}[i]\|_2} \cdot \frac{\mathbf{P}_{t,k}[j]^{\top}}{\|\mathbf{P}_{t,k}[j]\|_2} \right),
    \label{eq:distill_score}
\end{equation}
The final semantic-guided distillation coefficient is obtained by applying a temperature-scaled softmax over all accumulated teachers:
\begin{equation}
    \alpha_k =
    \frac{\exp(r_k / \tau)}
    {\sum_{m=0}^{t}\exp(r_m / \tau)}, \quad k=0,\ldots,t,
    \label{eq:distill_coeff}
\end{equation}
where $|\mathcal{Y}_k|$ is the number of categories in the $k$-th task, the former subscript in $\mathbf{P}_{i,j}$, stands for the task index, and the latter for the subspace, the $\mathbf{P}_{t,k}[j]$ is the prototype of the $j$-th class in the $t$-th task extracted by the $k$-th teacher. Specifically, the prototype is computed as:
\begin{equation}  \label{eq:alpha_prototype}
	\mathbf{P}_{t,k}[j]= \frac{1}{N} \sum_{i=1}^{|\mathcal{{D}}_t|}\mathbb{I}(y_i=j)\phi(\mathbf{x};\text{T}_k) ,
\end{equation}
where $\mathbb{I}$ denotes the indicator function, $N$ is the number of instances in class $j$, ${|\mathcal{{D}}_t|}$ is the number of examples in the current task, and $\tau$ is the temperature coefficient. The raw similarity scores $\{r_k\}_{k=0}^{t}$ are normalized via softmax to obtain the final distillation coefficients $\{\alpha_k\}_{k=0}^{t}$ used in Eq.~(\ref{eq:distill_loss}), ensuring that each teacher's contribution is proportional to its semantic relevance to the current task and that $\sum_{k=0}^{t}\alpha_k=1$.

{\color{black}The semantic coefficient $\alpha_k$ is a task-level quantity computed once per incremental stage from the current-task class prototypes. Current-task Teacher features are cached during coefficient construction, and the resulting coefficients are reused for all EKT mini-batches and epochs. It introduces no learnable parameters, optimizer state, or backward propagation. Given feature dimension $d$, the prototype-level similarity computation requires $\mathcal O(d|\mathcal Y_t|\sum_{k=0}^{t}|\mathcal Y_k|)$ scalar operations. This training-only computation does not affect deployment storage or inference computation.}

\subsection{Optimization Objective and Training Procedure}
\label{sec:optimization}
TALON employs a two-stage training paradigm that sequentially learns task-specific LoRA-Teacher modules and then consolidates their collective knowledge into a unified LoRA-Student through ensemble knowledge transfer. Algorithm~\ref{alg:talon} provides the complete training procedure.

\begin{algorithm}[!htbp]
    \caption{TALON for FSCIL}
    \label{alg:talon}
    \KwIn{Pre-trained model $\phi(\cdot)$, incremental datasets $\{\mathcal{D}_0, \ldots, \mathcal{D}_T\}$, LoRA rank $r$}
    \KwOut{LoRA-Student $\mathbf{S}_T$}
    \BlankLine
    Initialize Teacher storage list: $\mathcal{T} \leftarrow \emptyset$\;
    \For{$t = 0$ \KwTo $T$}{
        \tcp{\color{blue}Stage 1: Task-Adaptive LoRA-Teacher.}
        Initialize LoRA-Teacher $\mathbf{T}_t$: $\mathbf{V}_t = \mathbf{0}$, $\mathbf{U}_t \sim \text{Kaiming}$\;
        \textcolor{black}{Train $\mathbf T_t$ and the current-class columns of $\mathbf W_{\mathrm{head}}$ on $\mathcal D_t$ using Eq.~(\ref{eq:teacher_loss}); retain the historical columns unchanged\;}
        Freeze $\mathbf{T}_t$ parameters\;
        Append $\mathbf{T}_t$ to Teacher storage: $\mathcal{T} \leftarrow \mathcal{T} \cup \{\mathbf{T}_t\}$\;
        Extract prototypes $\{\mathbf{P}_{k,k}[j]\}$ using Eq.~(\ref{eq:alpha_prototype}) where $k = t$\;
        \BlankLine
        \tcp{\color{blue}Stage 2: Ensemble Knowledge Transfer.}
        \textcolor{black}{Freeze the complete shared linear head $\mathbf W_{\mathrm{head}}$ during EKT\;}
        \If{$t = 0$}{
            Initialize LoRA-Student $\mathbf{S}_0 = \mathbf{T}_0$\;
        }
        \Else{
            Update LoRA-Student $\mathbf{S}_t = \mathbf{S}_{t-1}$\;
        }
        \If{$t > 0$}{
        Compute semantic coefficients $\{\alpha_k\}_{k=0}^{t}$ over all Teachers in $\mathcal{T}$ via Eq.~(\ref{eq:distill_coeff})\;
        \For{mini-batch $(\mathbf{x}, \mathbf{y}) \in \mathcal{D}_t$}{
            Compute EKT loss $L_{\text{EKT}}$ using Eq.~(\ref{eq:total_distill_loss})\;
            Update $\mathbf{S}_t$ using $\nabla L_{\text{EKT}}$\;
        }}
        Extract the prototypes $\mathbf{P}_\text{Stu}[j]$ for classes $j \in \mathcal{Y}_t$ via Eq.~(\ref{eq:student_prototype})\;
        Append $\mathbf{P}_\text{Stu}$ to the prototype classifier; prototypes of $\mathcal{Y}_{<t}$ remain frozen\;
    }
    \textbf{return} the LoRA-Student $\mathbf{S}_T$\;
\end{algorithm}

\textbf{Stage 1: Task-Adaptive LoRA-Teacher.}
For each incremental task $t$, we train a dedicated LoRA-Teacher $\mathbf{T}_t$ while maintaining the pre-trained backbone in a frozen state. This stage optimizes the cross-entropy loss to ensure effective task-specific adaptation:
\begin{equation}
    \textcolor{black}{
    \min_{\mathbf{T}_t,\,\mathbf{W}_{\mathrm{head}}[:,\mathcal{Y}_t]}
    \mathbb{E}_{(\mathbf{x},\mathbf{y}) \in \mathcal{D}_t}
    \left[
    \text{L}_{\text{CE}}
    \left(
    \text{f}(\mathbf{x};\mathbf{T}_t),
    \mathbf{y}
    \right)
    \right]
    },
    \label{eq:teacher_loss}
\end{equation}
\textcolor{black}{where the pre-trained backbone remains frozen. At session $t$, the columns of $\mathbf W_{\mathrm{head}}$ corresponding to the current classes $\mathcal Y_t$ provide task-local supervision for the current LoRA-Teacher. Historical-class logits are masked when computing the current-task cross-entropy loss. During this stage, the current LoRA-Teacher and the current-class columns of $\mathbf W_{\mathrm{head}}$ are optimized. During the subsequent EKT stage, the complete shared head is frozen.}

\textbf{Stage 2: Ensemble Knowledge Transfer.}
Given the collection of trained LoRA-Teacher modules $\{\mathbf{T}_0, \ldots, \mathbf{T}_t\}$, we optimize a LoRA-Student $\mathbf{S}_t$ that consolidates their collective knowledge through the ensemble knowledge transfer mechanism described in Eq.~(\ref{eq:total_distill_loss}). This stage enables efficient single-model inference while preserving the benefits of multi-teacher expertise.

\textbf{Inference Protocol.}
{\color{black}
After EKT at session $t$, the current LoRA-Student $\mathbf{S}_t$ serves as the single deployed feature extractor. For each newly introduced class $c\in\mathcal{Y}_t$, its prototype is computed as
\begin{equation}
\mathbf{P}_\text{Stu}^{(t)}[c]=\frac{1}{|\mathcal{D}_t^c|}\sum_{(\mathbf{x}_i,y_i)\in\mathcal{D}_t^c}\phi(\mathbf{x}_i;\mathbf{S}_t),\qquad \mathcal{D}_t^c=\{(\mathbf{x}_i,y_i)\in\mathcal{D}_t:y_i=c\}.
\label{eq:student_prototype}
\end{equation}

The accumulated prototype bank is the class-indexed collection
\begin{equation}
\mathcal{P}_t=\left\{\mathbf{P}_\text{Stu}^{(j_c)}[c]:c\in\mathcal{Y}_{\leq t}\right\},
\label{eq:accumulated_prototype_bank}
\end{equation}
where $j_c$ denotes the session in which class $c$ was introduced. 
\textcolor{black}{At each session, prototypes of the newly introduced classes are appended to $\mathcal P_t$, whereas historical entries remain unchanged because TALON follows an exemplar-free protocol and does not retain historical samples. Consequently, a historical class $c$ introduced at session $j_c$ is represented by $\mathbf P_{\mathrm{Stu}}^{(j_c)}[c]$, while a test sample at a later session $t>j_c$ is represented using the current Student $\mathbf S_t$. Prediction is performed by}

\begin{equation}
{\color{black}
\hat y
=
\arg\max_{c\in\mathcal Y_{\leq t}}
\operatorname{sim}
\left(
\mathbf P_{\mathrm{Stu}}^{(j_c)}[c],
\phi(\mathbf x;\mathbf S_t)
\right),
}
\label{eq:prediction}
\end{equation}
\textcolor{black}{where $\operatorname{sim}(\cdot,\cdot)$ denotes cosine similarity. Because $\mathbf S_t$ continues to change while historical entries of $\mathcal P_t$ remain frozen, this protocol may introduce prototype--feature inconsistency across incremental stages. We quantify this effect in Section~\ref{sec:prototype_drift}.}

\textcolor{black}{TALON follows a strict exemplar-free protocol: no raw samples or mini-batches from previous tasks are stored or replayed. Before deployment, all LoRA-Teachers, their training-time prototype banks, and the shared linear head $\mathbf W_{\mathrm{head}}$ are discarded; inference requires only the current LoRA-Student $\mathbf S_t$ and the accumulated Student prototype bank $\mathcal P_t$. }

}
\section{Experiments}
\label{sec:experiments}

This section provides a comprehensive evaluation of TALON on multiple benchmarks.

\subsection{Implementation Details}

\textbf{Datasets:} We follow~\cite{liu2024few,liu2025sec,MTE-FSCIL} to evaluate the performance on four benchmark datasets: CUB200~\cite{wah2011caltech}, CIFAR100~\cite{krizhevsky2009learning}, ImageNet-R~\cite{hendrycks2021many}, and \textit{mini}ImageNet~\cite{russakovsky2015imagenet}. CUB200 and ImageNet-R contain 200 classes, whereas CIFAR100  and \textit{mini}ImageNet each include 100 classes. For all datasets, we adopt the split configuration proposed in~\cite{liu2024few}, as summarized in Table~\ref{tab:fscil-config}.

\begin{table*}[!htbp]
    \caption{Setups for the four datasets}
    \label{tab:fscil-config}
    \centering
    \begin{tabular*}{\textwidth}{@{\extracolsep{\fill}}lccccc@{}}
        \toprule
        Task  &CUB200  &CIFAR100  &ImageNet-R  &\textit{mini}ImageNet  \\
        \midrule
        Base &100  &60  &100  &60            \\
        Incremental    & 10-way 5-shot & 5-way 5-shot  & 10-way 5-shot  & 5-way 5-shot \\
        \# of tasks & 1+10  & 1+8   & 1+10  & 1+8   \\
        \bottomrule
    \end{tabular*}
\end{table*}

\textbf{Comparison methods:} We evaluate TALON against state-of-the-art approaches across two categories: PEFT-based CIL methods, including L2P~\cite{wang2022learning}, CODA-Prompt~\cite{smith2023coda}, LAE~\cite{gao2023unified}, InfLoRA~\cite{liang2024inflora}, and SD-LoRA~\cite{wu2025sdlora}; and PEFT-based FSCIL methods, including ASP~\cite{liu2024few} and SEC-prompt~\cite{liu2025sec}. For broader comparison, we additionally consider SimpleFSCIL~\cite{zhou2024revisiting}, which employs a prototype-based classifier on a frozen pre-trained backbone, as well as standard full fine-tuning. All methods are evaluated under identical conditions with the same pre-trained backbone (ViT-B/16-IN1K~\cite{dosovitskiy2020image}) and dataset splits.

\textbf{Evaluation metrics:} We evaluate model performance using five established metrics~\cite{park2024pre,tian2024survey}: $\mathcal{A}_{\text{Base}}$, $\mathcal{A}_L$, $\bar{\mathcal{A}}$, performance drop (PD, in pp)~\cite{liu2024few}, and forward transfer (FWT)~\cite{gem}. Specifically, $\mathcal{A}_{\text{Base}}$ denotes the accuracy on the base task, $\mathcal{A}_L$ denotes the accuracy on the final incremental task, and $\bar{\mathcal{A}} = \frac{1}{T}\sum_{i=1}^{T}\mathcal{ACC}_{i}$ measures the average accuracy across all $T$ incremental stages, where $\mathcal{ACC}_{i} = \frac{1}{i} \sum_{j=1}^{i}a_{i,j}$, with $a_{i,j}$ representing the accuracy on the $j$-th task after training on the $i$-th task~\cite{liang2024inflora}. 

The performance drop (PD) quantifies knowledge forgetting, i.e., the absolute accuracy decline (in percentage points) from the base to the final session: $\text{PD} = \mathcal{A}_{\text{Base}} - \mathcal{A}_L$, where lower values indicate stronger retention of learned knowledge.

Forward transfer (FWT) evaluates the influence of learning task $t$ on the performance of a future task $k > t$, capturing the model’s ability to generalize to unseen tasks in a zero-shot manner~\cite{gem}: $\text{FWT} = \frac{1}{T-1} \sum_{i=2}^{T} \left( a_{i-1, i} - a_{\text{rand}, i} \right)$,
where $a_{i-1, i}$ represents the accuracy on task $i$ after training on task $i-1$, and $a_{\text{rand}, i}$ is the accuracy on task $i$ using a randomly initialized model with prototype-based classification computed from the incremental training data. Positive FWT indicates beneficial knowledge transfer across tasks. When comparing models with similar $\bar{\mathcal{A}}$, the one with higher FWT is preferred. Note that FWT is not computed for the final task as there are no subsequent tasks for evaluation.

\textbf{Training details:} 
We employ ViT-B/16-IN1K~\cite{dosovitskiy2020image} as the backbone architecture. For optimization, we utilize SGD with learning rates of 0.01 for MLP-Teachers and 0.02 for QV-Teachers, employing cosine annealing learning rate decay. LoRA-Teachers are trained for 20 epochs on the base task and 10 epochs for each incremental task, with batch sizes of 48 and 16, respectively. We set the LoRA rank to $r=8$. \textcolor{black}{For EKT, we use 5 distillation epochs with batch size 16, semantic temperature $\tau=1$, and fixed equal weights for $L_{\mathrm{SD}}$ and $L_{\mathrm{HD}}$. The supplementary analysis shows limited sensitivity to $\tau$, with the four-dataset macro-average accuracy varying by only 0.12 percentage points over $\tau\in\{0.25,0.5,1,2,4\}$.} \textcolor{black}{Unless otherwise specified, experiments use seed 1993; multi-seed experiments use seeds 42, 1993, and 2025.}

\textbf{TALON Variants:} We evaluate two variants: \textbf{TALON-QV} tunes Query and Value projections, consistent with existing LoRA-based methods (InfLoRA~\cite{liang2024inflora}, SD-LoRA~\cite{wu2025sdlora}), enabling fair comparison; \textbf{TALON-MLP} tunes MLP layers, achieving comparable performance with 2$\times$ fewer parameters (0.15M vs. 0.29M). This demonstrates that our framework is agnostic to the specific PEFT location.

\subsection{Benchmark Comparison}

\begin{figure*}
    \centering
    \captionsetup[subfloat]{font=scriptsize,skip=2pt}
    \subfloat[CUB200 ($T{=}11$)\label{fig:fscil_benchmarks_curves_cub}]{\makebox[0.25\linewidth][c]{\includegraphics[width=0.235\linewidth]{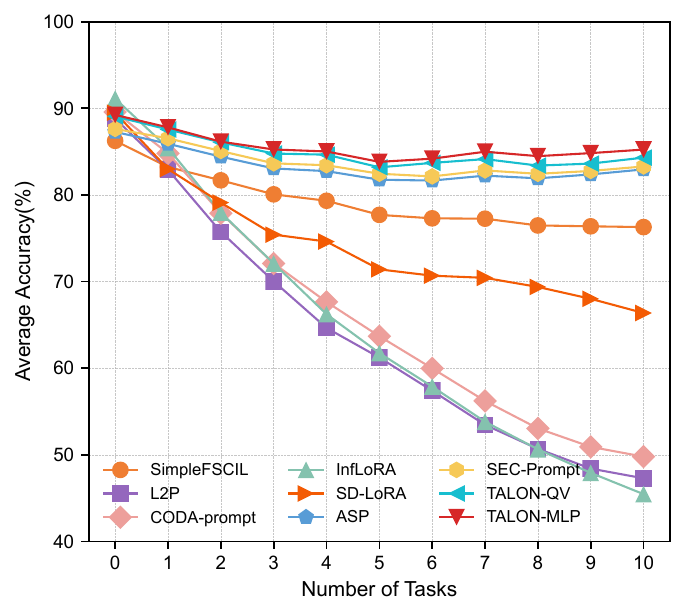}}}%
    \subfloat[CIFAR100 ($T{=}9$)\label{fig:fscil_benchmarks_curves_cifar}]{\makebox[0.25\linewidth][c]{\includegraphics[width=0.235\linewidth]{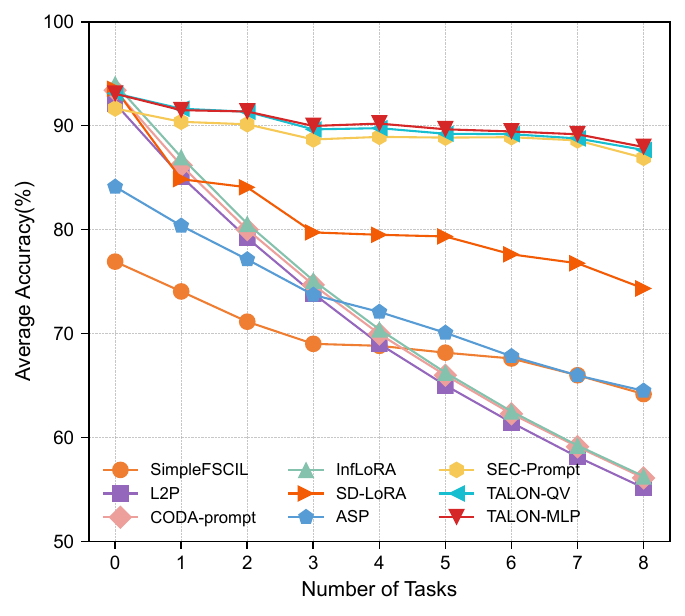}}}%
    \subfloat[ImageNet-R ($T{=}11$)\label{fig:fscil_benchmarks_curves_imagenetr}]{\makebox[0.25\linewidth][c]{\includegraphics[width=0.235\linewidth]{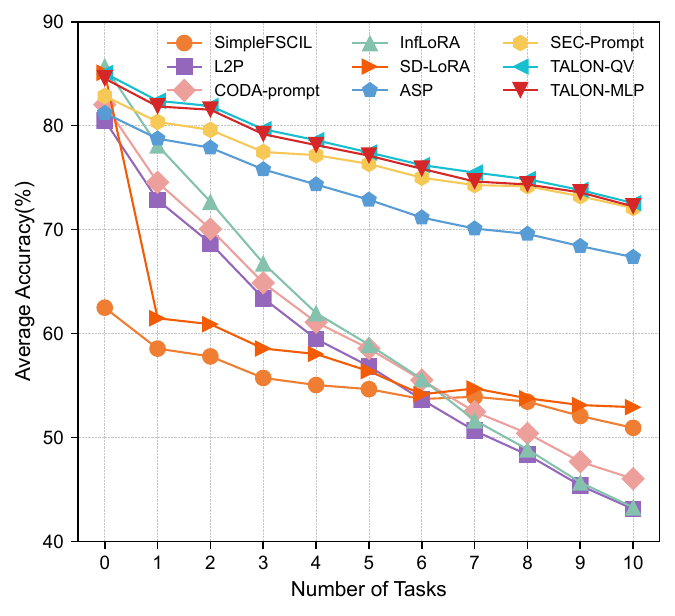}}}%
    \subfloat[\textit{mini}ImageNet ($T{=}9$)\label{fig:fscil_benchmarks_curves_mini}]{\makebox[0.25\linewidth][c]{\includegraphics[width=0.235\linewidth]{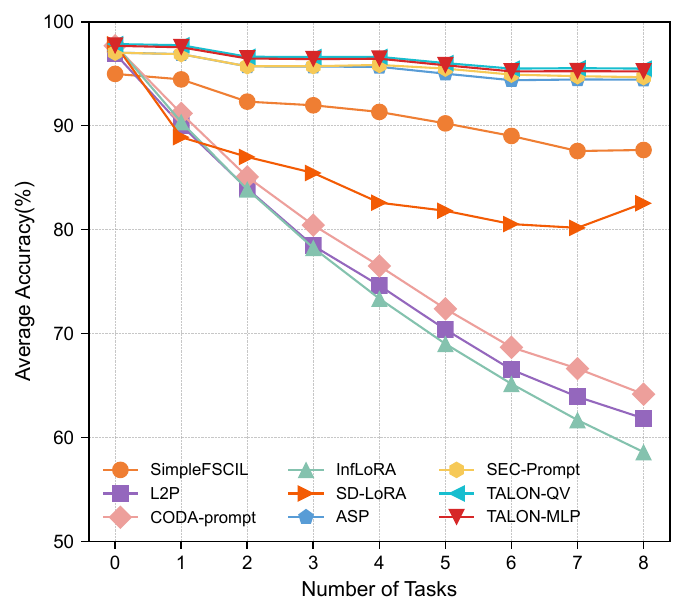}}}
    \caption{Performance curves on different datasets. All methods are based on the same pre-trained model ({\bf ViT-B/16-IN1K}).}
    \label{fig:fscil_benchmarks_curves}
\end{figure*}

We conduct a comprehensive evaluation of TALON against state-of-the-art methods on four benchmark datasets: CUB200, CIFAR100, ImageNet-R, and \textit{mini}ImageNet. Results are reported in Table~\ref{tab:fscil_benchmarks_acc} and Table~\ref{tab:fscil_benchmarks_fwt}. Performance trajectories across incremental tasks are further illustrated in Fig.~\ref{fig:fscil_benchmarks_curves}.

\begin{table*}[t]
        \centering
        \color{black}
        \scriptsize
        \setlength{\tabcolsep}{1.2pt}
        \caption{\textcolor{black}{Comparison with state-of-the-art FSCIL methods. Values are mean$\pm$sample standard deviation over three class-order runs generated using seeds 42, 1993, and 2025. All methods use the same ViT-B/16-IN1K backbone and dataset splits. The best mean is highlighted in \textbf{bold}, and the second-best mean is \underline{underlined}.}}
       \label{tab:fscil_benchmarks_acc}
        \resizebox{\linewidth}{!}{
        \begin{tabular}{@{}lcccccccccccc@{}}
            \toprule
            Method
            & \multicolumn{3}{c}{CUB200 ($T=11$)}
            & \multicolumn{3}{c}{CIFAR100 ($T=9$)}
            & \multicolumn{3}{c}{ImageNet-R ($T=11$)}
            & \multicolumn{3}{c}{\textit{mini}ImageNet ($T=9$)}
            \\
            \cmidrule(lr){2-4}
            \cmidrule(lr){5-7}
            \cmidrule(lr){8-10}
            \cmidrule(l){11-13}
            & $\mathcal{A}_{\mathrm{Base}}$
            & $\mathcal{A}_L$
            & $\bar{\mathcal{A}}$
            & $\mathcal{A}_{\mathrm{Base}}$
            & $\mathcal{A}_L$
            & $\bar{\mathcal{A}}$
            & $\mathcal{A}_{\mathrm{Base}}$
            & $\mathcal{A}_L$
            & $\bar{\mathcal{A}}$
            & $\mathcal{A}_{\mathrm{Base}}$
            & $\mathcal{A}_L$
            & $\bar{\mathcal{A}}$
            \\
            \midrule

            Full Finetune
            & \mstd{89.36}{0.44}
            & \mstd{11.28}{3.09}
            & \mstd{22.19}{6.46}
            & \mstd{92.08}{0.54}
            & \mstd{40.86}{7.14}
            & \mstd{66.17}{4.02}
            & \mstd{82.66}{0.76}
            & \mstd{13.23}{2.93}
            & \mstd{27.99}{7.29}
            & \mstd{95.60}{0.52}
            & \mstd{69.04}{4.67}
            & \mstd{80.88}{1.67}
            \\

            SimpleFSCIL
            & \mstd{88.57}{2.01}
            & \mstd{74.67}{1.41}
            & \mstd{79.71}{0.90}
            & \mstd{78.13}{1.51}
            & \mstd{63.85}{0.55}
            & \mstd{69.75}{0.69}
            & \mstd{63.73}{1.10}
            & \mstd{50.32}{0.79}
            & \mstd{55.46}{0.64}
            & \mstd{94.74}{0.66}
            & \mstd{87.55}{1.07}
            & \mstd{90.65}{0.47}
            \\

            L2P
            & \mstd{90.41}{1.89}
            & \mstd{48.33}{1.06}
            & \mstd{65.13}{1.26}
            & \mstd{92.43}{0.72}
            & \mstd{55.43}{0.27}
            & \mstd{71.25}{0.38}
            & \mstd{79.79}{0.96}
            & \mstd{42.37}{2.03}
            & \mstd{57.51}{1.67}
            & \mstd{97.01}{0.30}
            & \mstd{62.09}{0.52}
            & \mstd{77.05}{0.67}
            \\

            CODA-Prompt
            & \mstd{91.34}{1.92}
            & \mstd{50.38}{0.88}
            & \mstd{66.75}{0.68}
            & \mstd{93.50}{0.26}
            & \mstd{56.25}{0.13}
            & \mstd{72.12}{0.14}
            & \mstd{81.70}{0.36}
            & \mstd{45.67}{1.45}
            & \mstd{60.01}{1.76}
            & \mstd{97.65}{0.14}
            & \mstd{63.46}{0.91}
            & \mstd{77.68}{0.47}
            \\

            LAE
            & \mstd{90.91}{2.37}
            & \mstd{50.53}{3.99}
            & \mstd{66.67}{2.85}
            & \mstd{92.19}{1.44}
            & \mstd{55.97}{2.14}
            & \mstd{71.51}{1.79}
            & \mstd{76.89}{2.89}
            & \mstd{42.79}{5.41}
            & \mstd{56.44}{4.59}
            & \mstd{97.09}{0.40}
            & \mstd{64.34}{5.56}
            & \mstd{78.29}{3.08}
            \\

            InfLoRA
            & \textbf{\mstd{92.24}{1.02}}
            & \mstd{45.36}{0.58}
            & \mstd{64.73}{0.31}
            & \textbf{\mstd{94.43}{0.46}}
            & \mstd{56.47}{0.22}
            & \mstd{72.49}{0.24}
            & \textbf{\mstd{85.05}{0.59}}
            & \mstd{43.37}{2.10}
            & \mstd{60.22}{1.92}
            & \underline{\mstd{97.84}{0.09}}
            & \mstd{58.69}{0.12}
            & \mstd{75.41}{0.09}
            \\

            SD-LoRA
            & \underline{\mstd{91.51}{1.91}}
            & \mstd{60.05}{6.01}
            & \mstd{70.73}{3.78}
            & \underline{\mstd{94.16}{0.84}}
            & \mstd{74.37}{0.80}
            & \mstd{81.71}{0.64}
            & \mstd{84.17}{0.81}
            & \mstd{51.65}{1.15}
            & \mstd{57.91}{1.23}
            & \textbf{\mstd{97.88}{0.22}}
            & \mstd{80.56}{4.39}
            & \mstd{84.79}{1.65}
            \\

            ASP
            & \mstd{90.36}{2.69}
            & \mstd{82.78}{0.15}
            & \mstd{85.73}{2.23}
            & \mstd{89.89}{5.04}
            & \mstd{78.79}{12.39}
            & \mstd{83.72}{9.40}
            & \mstd{81.82}{1.10}
            & \mstd{70.75}{3.28}
            & \mstd{75.40}{2.32}
            & \mstd{96.85}{0.22}
            & \mstd{94.38}{0.49}
            & \mstd{95.33}{0.51}
            \\

            SEC-prompt
            & \mstd{90.85}{2.84}
            & \underline{\mstd{83.56}{0.24}}
            & \underline{\mstd{86.49}{2.43}}
            & \mstd{92.42}{0.74}
            & \mstd{87.02}{0.13}
            & \mstd{89.53}{0.27}
            & \mstd{83.19}{0.35}
            & \underline{\mstd{73.64}{1.36}}
            & \mstd{77.57}{0.98}
            & \mstd{96.72}{0.28}
            & \mstd{94.66}{0.14}
            & \mstd{95.46}{0.25}
            \\

            \midrule

            TALON-MLP
            & \mstd{91.41}{1.89}
            & \textbf{\mstd{83.94}{1.15}}
            & \textbf{\mstd{86.68}{1.22}}
            & \mstd{93.68}{0.68}
            & \textbf{\mstd{87.56}{0.63}}
            & \textbf{\mstd{90.39}{0.27}}
            & \mstd{84.85}{0.59}
            & \mstd{73.57}{1.22}
            & \underline{\mstd{78.19}{1.01}}
            & \mstd{97.61}{0.12}
            & \underline{\mstd{95.30}{0.32}}
            & \underline{\mstd{96.16}{0.33}}
            \\

            TALON-QV
            & \mstd{91.45}{2.16}
            & \mstd{83.42}{0.78}
            & \mstd{86.39}{1.55}
            & \mstd{93.69}{0.60}
            & \underline{\mstd{87.41}{0.38}}
            & \underline{\mstd{90.29}{0.25}}
            & \underline{\mstd{85.03}{0.55}}
            & \textbf{\mstd{73.71}{1.18}}
            & \textbf{\mstd{78.38}{0.94}}
            & \mstd{97.63}{0.23}
            & \textbf{\mstd{95.56}{0.35}}
            & \textbf{\mstd{96.34}{0.33}}
            \\

            \bottomrule
        \end{tabular}}
\end{table*}

\begin{table*}[!htbp]
    \centering
    \caption{\textcolor{black}{Comparison with state-of-the-art FSCIL methods under seed 1993. We report performance drop (PD, in pp) and forward transfer (FWT) using the same ViT-B/16-IN1K backbone. }}
    \label{tab:fscil_benchmarks_fwt}
    \setlength{\tabcolsep}{10pt}
    \resizebox{1.0\linewidth}{!}{%
        \begin{tabular}{@{}lcccccccc@{}}
            \toprule
            \multirow{2}{*}{Method} & 
            \multicolumn{2}{c}{CUB200 ($T$=11)} & 
            \multicolumn{2}{c}{CIFAR100 ($T$=9)} &
            \multicolumn{2}{c}{ImageNet-R ($T$=11)} &
            \multicolumn{2}{c}{\textit{mini}ImageNet ($T$=9)}
            \\
            \cmidrule(lr){2-3} \cmidrule(l){4-5} \cmidrule(l){6-7} \cmidrule(l){8-9}
            &PD$\downarrow$ & FWT$\uparrow$
            &PD$\downarrow$ & FWT$\uparrow$
            &PD$\downarrow$ & FWT$\uparrow$
            &PD$\downarrow$ & FWT$\uparrow$\\
            \midrule
        Full Finetune 
        &79.49 &-17.13
        &51.52 &34.80
        &66.60 &15.69
        &31.92 &9.02
        \\
        
        SimpleFSCIL 
        &9.96 &13.67
        &12.74 &33.80
        &\underline{11.56} &35.89
        &7.32 &-0.47
        \\
        
        L2P 
        &41.52 &11.67
        &36.97 &40.80
        &37.37 &38.89
        &35.08 &15.52
        \\
        
        CODA-Prompt
        &39.79 &\underline{17.78}
        &37.29 &44.30
        &36.00 &\textbf{45.29}
        &33.51 &13.02
        \\
        
        LAE 
        &42.16 &9.07
        &36.83 &45.30
        &36.34 &35.49
        &38.56 &-30.98
        \\

        InfLoRA
        &45.66 &14.87
        &37.75 &37.30
    &42.44 &\underline{44.89}
        &39.18 &\textbf{17.02}
        \\

        SD-LoRA
    &23.13 &\textbf{23.60}
    &19.21 &\underline{47.50}
        &32.17 &35.01
        &15.28 &13.92
        \\
        ASP
    &\underline{4.33} &9.47
        &19.63 &12.30
        &13.86 &17.69
        &2.60 &6.52
        \\
        SEC-prompt
        &\underline{4.33} &9.47
        &\textbf{4.75} &30.80
        &\textbf{10.79} &26.09
        &\textbf{2.36} &8.52
        \\
        \midrule
        TALON-MLP 
        &\textbf{4.00} &13.27
        &\underline{5.12} &39.30
        &12.30 &39.49
        &\underline{2.46} &16.02
        \\
                TALON-QV 
                &4.76 &16.67
                &5.44 &\textbf{47.80}
                &12.51 &41.49
            &\textbf{2.36} &\underline{16.52}

        \\
        \bottomrule
                \end{tabular}}
\end{table*}

\textbf{Accuracy:} 
\textcolor{black}{As shown in Table~\ref{tab:fscil_benchmarks_acc}, TALON \textcolor{black}{obtains comparable or better} mean last-session and average accuracy \textcolor{black}{across} all four benchmarks. On CIFAR100, ImageNet-R, and \textit{mini}ImageNet, the best TALON variant exceeds SEC-prompt in mean average accuracy by 0.86, 0.81, and 0.88 percentage points, respectively. On CUB200, TALON-MLP and SEC-prompt obtain comparable mean average accuracies of $86.68\pm1.22\%$ and $86.49\pm2.43\%$, respectively, and their relative ranking varies across class orders.} These \textcolor{black}{results} highlight TALON’s ability to effectively balance stability and plasticity in FSCIL scenarios.


\textbf{Forgetting:} Table~\ref{tab:fscil_benchmarks_fwt} shows that conventional PEFT-based methods such as InfLoRA and SD-LoRA suffer from severe catastrophic forgetting. The limited training samples in FSCIL provide insufficient gradient signals, leading to both overfitting to new classes and erosion of previously learned knowledge. \textcolor{black}{By contrast, TALON maintains relatively low PD values across all four datasets.} For example, on CUB200, TALON-MLP obtains the lowest PD of 4.00\%, substantially lower than InfLoRA (45.66\%) and SD-LoRA (23.13\%). \textcolor{black}{Across CIFAR100, ImageNet-R, and \textit{mini}ImageNet, TALON maintains competitive knowledge retention, while SEC-prompt obtains lower or equal PD values.}


\textbf{Forward Transfer:} 
\textcolor{black}{TALON also exhibits positive forward transfer (FWT) across all datasets (Table~\ref{tab:fscil_benchmarks_fwt}).} Compared with methods achieving similar accuracy (e.g., SEC-prompt and ASP), TALON demonstrates a stronger ability to leverage prior knowledge for learning novel tasks. For instance, on CUB200, TALON-QV achieves an FWT of 16.67\%, significantly higher than SEC-prompt’s 9.47\%. On CIFAR100, TALON-QV obtains the best FWT at 47.80\%, markedly surpassing SEC-prompt’s 30.80\%. On \textit{mini}ImageNet, TALON-QV records 16.52\%, outperforming SEC-prompt’s 8.52\%. This indicates TALON’s capacity for positive knowledge transfer, which is particularly valuable in few-shot incremental settings.

Following common FSCIL reporting practice, we additionally provide detailed per-session Top-1 accuracies on CUB200 in Table~\ref{tab:stage_acc_cub_seed1993}, which verifies that TALON maintains consistently strong performance throughout the incremental sequence rather than only at the final session.

\begin{table*}[t]
    \centering
    \caption{Detailed Top-1 accuracy (\%) at each FSCIL session on CUB200 under seed 1993. We report per-session accuracy $A_t$, average accuracy $\bar{\mathcal{A}}$, and performance drop (PD, in pp).}
    \label{tab:stage_acc_cub_seed1993}
    \setlength{\tabcolsep}{1.5pt}
    \resizebox{\linewidth}{!}{%
    \begin{tabular}{@{}lccccccccccccc@{}}
        \toprule
        Method & $A_0$ & $A_1$ & $A_2$ & $A_3$ & $A_4$ & $A_5$ & $A_6$ & $A_7$ & $A_8$ & $A_9$ & $A_{10}$ & $\bar{\mathcal{A}}$ & PD$\downarrow$ \\
        \midrule
        Full Finetune & 88.90 & 2.28 & 3.30 & 7.91 & 6.18 & 9.83 & 10.49 & 13.00 & 11.88 & 7.70 & 9.41 & 15.53 & 79.49 \\
        SimpleFSCIL & 86.25 & 83.23 & 81.69 & 80.07 & 79.33 & 77.70 & 77.30 & 77.26 & 76.48 & 76.38 & 76.29 & 79.27 & 9.96 \\
        L2P & 88.81 & 82.91 & 75.74 & 70.03 & 64.65 & 61.23 & 57.40 & 53.46 & 50.69 & 48.41 & 47.29 & 63.69 & 41.52 \\
        CODA-Prompt & 89.58 & 84.80 & 77.89 & 72.09 & 67.65 & 63.69 & 59.98 & 56.22 & 53.05 & 50.92 & 49.79 & 65.97 & 39.79 \\
        LAE & 88.56 & 82.28 & 75.09 & 69.50 & 65.02 & 61.01 & 57.45 & 53.61 & 50.54 & 48.50 & 46.40 & 63.45 & 42.16 \\
        InfLoRA & 91.12 & 85.43 & 77.96 & 72.09 & 66.24 & 61.75 & 57.88 & 53.82 & 50.64 & 47.92 & 45.46 & 64.57 & 45.66 \\
        SD-LoRA & 89.50 & 82.99 & 79.11 & 75.42 & 74.62 & 71.41 & 70.68 & 70.43 & 69.38 & 68.01 & 66.37 & 74.36 & 23.13 \\
        ASP & 87.28 & 85.91 & 84.42 & 83.06 & 82.75 & 81.76 & 81.66 & 82.23 & 81.92 & 82.37 & 82.95 & 83.30 & 4.33 \\
        SEC-prompt & 87.62 & 86.54 & 85.07 & 83.65 & 83.43 & 82.45 & 82.14 & 82.83 & 82.44 & 82.77 & 83.29 & 83.84 & 4.33 \\
        \midrule
        TALON-MLP & 89.24 & 87.80 & 86.15 & 85.25 & 85.02 & 83.82 & 84.19 & 84.99 & 84.48 & 84.83 & 85.24 & 85.55 & 4.00 \\
        TALON-QV & 89.07 & 87.56 & 86.07 & 84.78 & 84.65 & 83.19 & 83.70 & 84.14 & 83.39 & 83.62 & 84.31 & 84.95 & 4.76 \\
        \bottomrule
    \end{tabular}}
\end{table*}

Finally, the performance curves in Fig.~\ref{fig:fscil_benchmarks_curves} confirm that TALON maintains consistently high accuracy throughout the entire incremental learning process, underscoring its robustness and stability. 

Longer task sequences ($T$=21) and a larger backbone (ViT-L/16) are further evaluated in the supplementary material, where TALON's advantages remain consistent.

\subsection{Efficiency Analysis} 


\textcolor{black}{This subsection evaluates computational efficiency by explicitly separating training-phase and deployment-phase costs. Deployment efficiency is measured using inference latency and the model state retained for inference. Training efficiency is measured using training-state parameters, peak GPU memory, per-epoch training time, cumulative Teacher-training time, cumulative EKT time, and end-to-end wall-clock time. We first compare TALON with baselines under the standard FSCIL sequences and then evaluate TALON-QV under the longer $T=21$ setting.}

\textbf{Inference Efficiency:} 
Many existing methods incur substantial computational overhead during inference due to reliance on dynamic module selection or runtime module generation. Specifically, prompt-based continual learning approaches such as L2P and CODA-Prompt require additional inference time for prompt selection on each input sample. LAE introduces extra forward passes for ensemble operations, while ASP and SEC-prompt employ computationally intensive encoder networks to generate task-specific modules conditioned on input features.

\begin{figure}[t]
    \begin{center}
        \includegraphics[width=0.97\columnwidth]{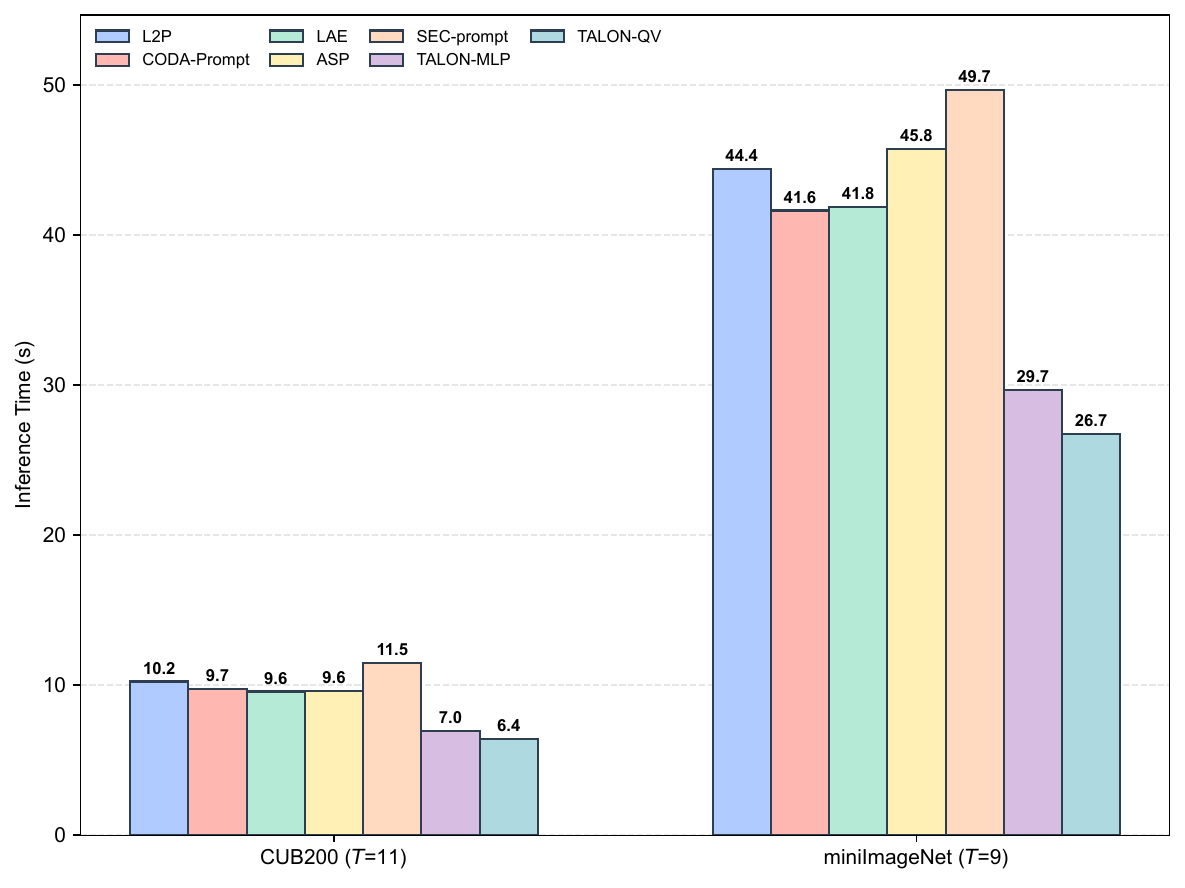}
    \end{center}
    \caption{Inference time per task (seconds) for different methods. Here, ``per task'' denotes one FSCIL session, and each value measures the wall-clock evaluation time at task $t$ on all classes seen so far. Experiments are conducted on an NVIDIA A800 GPU, with batch size 48 for the base task and 16 for incremental tasks.}
    \label{fig:inference_time}
\end{figure}

In contrast, TALON eliminates these bottlenecks via its Ensemble Knowledge Transfer (EKT) mechanism, which consolidates knowledge from all LoRA-Teachers into a single LoRA-Student during training. This design enables efficient single-model inference without runtime module selection or additional ensemble computations. Fig.~\ref{fig:inference_time} presents a comprehensive comparison of inference times across methods, demonstrating that TALON achieves the lowest latency for all evaluated tasks. For example, TALON-QV completes inference for a single task in \textcolor{black}{26.7} s, approximately \textcolor{black}{46.3\%} faster than SEC-prompt (49.7 s) and \textcolor{black}{\textbf{41.7\%}} faster than ASP (\textcolor{black}{45.8~s}). This substantial reduction in inference time highlights TALON’s suitability for real-time deployment in resource-constrained scenarios.

\begin{figure}
    \centering
    \subfloat[CUB200 ($T$=11)\label{fig:expanded-parameters-cub}]{\includegraphics[width=0.32\linewidth]{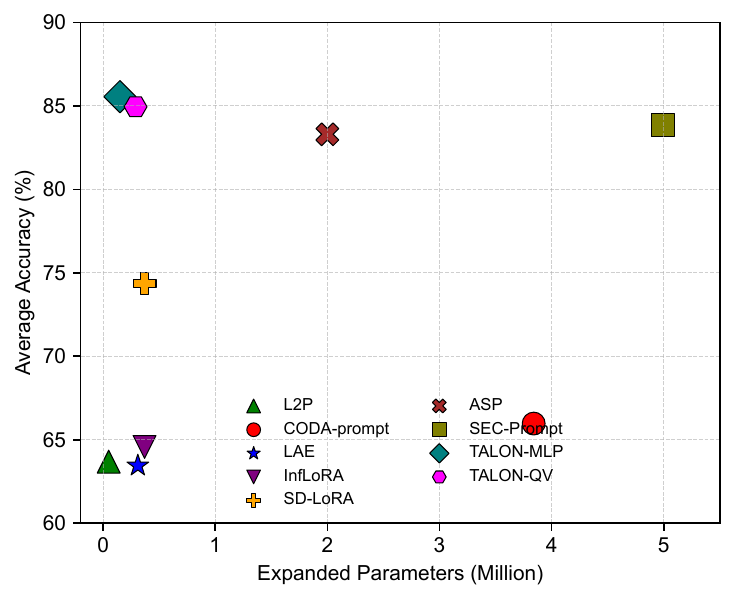}}%
    \hfil
    \subfloat[CIFAR100 ($T$=9)\label{fig:expanded-parameters-cifar}]{\includegraphics[width=0.32\linewidth]{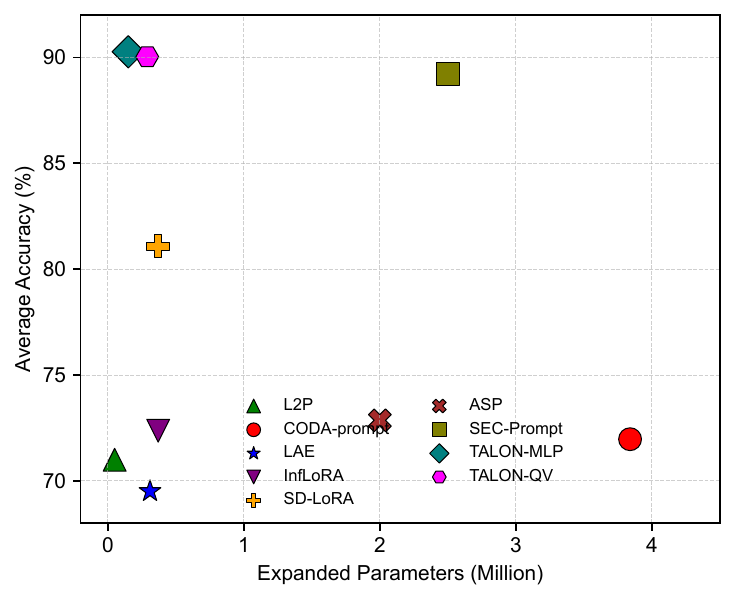}}%
    \subfloat[\textit{mini}ImageNet ($T$=9)\label{fig:expanded-parameters-mini}]{\includegraphics[width=0.32\linewidth]{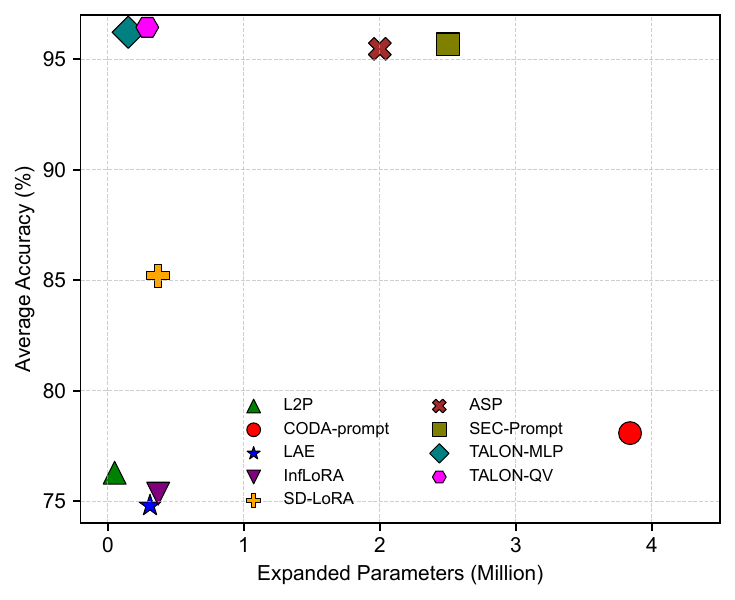}}
    \caption{The number of expanded parameters and accuracy of different methods.}
    \label{fig:expanded-parameters}
\end{figure}

\textbf{Expanded Parameters Analysis:} 
An important distinction in TALON is between training-phase and deployment-phase expanded parameters. \textcolor{black}{During training, TALON retains all accumulated LoRA-Teachers, the LoRA-Student, the auxiliary classifier, and the Teacher and Student prototype banks. Before deployment, the LoRA-Teachers, Teacher prototype banks, and auxiliary classifier are discarded; inference retains only the LoRA-Student and the Student prototype classifier. \textcolor{black}{Accordingly, Fig.~\ref{fig:expanded-parameters} compares deployment-phase expanded parameters and accuracy on CUB200, CIFAR100, and \textit{mini}ImageNet, where expanded parameters refer to the additional learned model parameters retained for inference beyond the frozen pre-trained backbone.}}

The parameter composition of each method is as follows: L2P and CODA-Prompt add parameters via learnable prompts and their keys; LAE incorporates LoRA modules alongside ensemble components; ASP and SEC-prompt employ both task-specific and task-agnostic prompts. In contrast, TALON’s expanded parameters are solely attributed to the LoRA-Student model, maintaining a parameter count comparable to InfLoRA and SD-LoRA while delivering superior accuracy. 

Experimental results demonstrate that TALON establishes a new benchmark in the parameter-performance trade-off. Both TALON-MLP and TALON-QV achieve superior accuracy with substantially fewer parameters than existing methods. For example, on CIFAR100, TALON-MLP attains 90.26\% accuracy with only 0.15M parameters, outperforming SEC-prompt (89.22\% with 2.50M parameters) while using over 16× fewer parameters. On CUB200, TALON-MLP achieves 85.55\%, surpassing SEC-prompt (83.84\%) with approximately \textbf{33×} fewer parameters (0.15M vs. 4.99M).

Notably, TALON maintains consistent parameter efficiency across datasets, unlike methods such as SEC-prompt, whose parameter count varies with dataset size. When compared to other LoRA-based approaches with similar budgets, e.g., InfLoRA (0.37M) and SD-LoRA (0.37M), TALON’s advantages become even more pronounced. On \textit{mini}ImageNet, TALON-QV (0.29M parameters) achieves 96.44\%, substantially higher than InfLoRA (75.34\%) and SD-LoRA (85.20\%). These results highlight TALON’s exceptional parameter-performance trade-off, demonstrating its ability to achieve high accuracy while remaining computationally efficient—particularly valuable in resource-constrained continual learning settings.

{\color{black}
\textbf{Training Efficiency:}
\label{sec:training_efficiency}
At session $t$, TALON first trains a new task-specific LoRA-Teacher and then distills all $t+1$ accumulated frozen Teachers into the LoRA-Student. Because the Teachers require forward computation but no gradient computation during EKT, the Teacher-side cost of each EKT epoch grows as $O(t)$.

Excluding the shared frozen pre-trained model, the retained training state at session $t$ is
\begin{equation}
M_{\mathrm{train}}(t)=(t+1)P_{\mathrm T}+P_{\mathrm S}+P_{\mathrm{aux}}+M_{\mathrm{proto}}(t),
\label{eq:training_state}
\end{equation}
where $P_{\mathrm T}$, $P_{\mathrm S}$, and $P_{\mathrm{aux}}$ denote one LoRA-Teacher, the LoRA-Student, and the auxiliary classifier, respectively, while $M_{\mathrm{proto}}(t)$ denotes the retained Teacher and Student prototype banks.

After training, the accumulated Teachers, Teacher prototype banks, and auxiliary classifier are not required for inference and can be discarded. The deployment state is therefore
\begin{equation}
M_{\mathrm{deploy}}=P_{\mathrm S}+C_{\leq T}d,
\label{eq:deployment_state}
\end{equation}
where $C_{\leq T}$ is the number of observed classes and $d$ is the feature dimension. Thus, the number of accumulated Teachers affects training-phase computation and retained training state, while deployment retains one LoRA-Student and one Student prototype classifier.
}

\begin{table}[t!]
\centering
\color{black}
\footnotesize
\setlength{\tabcolsep}{4pt}
\caption{\textcolor{black}{Performance and efficiency of TALON-QV under the dataset-specific $T=11/T=9$ settings and the extended $T=21$ setting. Training times are mean$\pm$sample standard deviation over three repetitions (seeds 42, 1993, and 2025) on the same NVIDIA A800; peak GPU memory and state sizes are deterministic. Retained training state excludes the frozen PTM and optimizer state, while deployment retains only the LoRA-Student and Student prototype classifier.}}
\label{tab:talon_qv_training_scalability_main}

\begin{tabular*}{\textwidth}{@{\extracolsep{\fill}}llcccc@{}}
\toprule
\multicolumn{6}{c}{\textbf{(a) Accuracy and cumulative time}} \\
\midrule
Dataset & $T$ & $\bar{\mathcal A}$ (\%) & \shortstack{Teacher\\train (s)} & \shortstack{EKT\\time (s)} & \shortstack{Total\\time (s)} \\
\midrule
CUB200 & $T=11$ & 84.95 & $650.3\pm2.2$ & $100.2\pm0.3$ & $787.8\pm1.6$ \\
CUB200 & $T=21$ & 85.24 & $689.8\pm4.1$ & $199.5\pm4.4$ & $932.4\pm8.4$ \\
\midrule
CIFAR100 & $T=9$ & 90.03 & $3553.0\pm63.3$ & $63.6\pm21.2$ & $3817.2\pm99.3$ \\
CIFAR100 & $T=21$ & 90.10 & $3613.0\pm85.4$ & $159.9\pm57.5$ & $3981.6\pm159.3$ \\
\midrule
ImageNet-R & $T=11$ & 77.99 & $1613.4\pm25.0$ & $118.9\pm24.1$ & $1826.2\pm46.6$ \\
ImageNet-R & $T=21$ & 77.83 & $1705.6\pm17.1$ & $276.7\pm57.6$ & $2094.2\pm83.5$ \\
\midrule
\textit{mini}ImageNet & $T=9$ & 96.44 & $4243.2\pm1126.1$ & $63.9\pm15.9$ & $4598.8\pm1255.0$ \\
\textit{mini}ImageNet & $T=21$ & 96.47 & $4657.0\pm969.9$ & $138.1\pm36.9$ & $5096.6\pm1056.7$ \\
\bottomrule
\end{tabular*}

\vspace{0.6em}

\begin{tabular*}{\textwidth}{@{\extracolsep{\fill}}llccc@{}}
\toprule
\multicolumn{5}{c}{\textbf{(b) Memory and retained states}} \\
\midrule
Dataset & $T$ & \shortstack{Peak GPU\\(MiB)} & \shortstack{Training state\\(MiB)} & \shortstack{Deploy. state\\(MiB)} \\
\midrule
CUB200 & $T=11$ & 4962.3 & 15.26 & 1.71 \\
CUB200 & $T=21$ & 4962.3 & 26.51 & 1.71 \\
\midrule
CIFAR100 & $T=9$ & 4961.0 & 12.13 & 1.42 \\
CIFAR100 & $T=21$ & 4961.0 & 25.63 & 1.42 \\
\midrule
ImageNet-R & $T=11$ & 4962.3 & 15.26 & 1.71 \\
ImageNet-R & $T=21$ & 4962.3 & 26.51 & 1.71 \\
\midrule
\textit{mini}ImageNet & $T=9$ & 4961.0 & 12.13 & 1.42 \\
\textit{mini}ImageNet & $T=21$ & 4961.0 & 25.63 & 1.42 \\
\bottomrule
\end{tabular*}
\end{table}

{\color{black}
\textbf{Long-Sequence Training Scalability:}
Using TALON-QV throughout, we compare $T=11$ with $T=21$ on CUB200 and ImageNet-R, and $T=9$ with $T=21$ on CIFAR100 and \textit{mini}ImageNet, where $T$ denotes the total number of sessions including the base session. Accuracy is evaluated using the protocol adopted in the main benchmark comparison. Efficiency statistics are measured on the same NVIDIA A800 and reported as mean$\pm$sample standard deviation over three complete timing repetitions. Teacher-training time is accumulated over all task-specific Teacher updates, cumulative EKT includes semantic-score computation and Student distillation, and total wall-clock time is measured end to end by the incremental training pipeline.
}

\textcolor{black}{Table~\ref{tab:talon_qv_training_scalability_main} summarizes TALON-QV performance and efficiency at $T=11/T=9$ and $T=21$. Extending the task sequence changes average accuracy by $+0.29$, $+0.07$, $-0.16$, and $+0.03$ percentage points on CUB200, CIFAR100, ImageNet-R, and \textit{mini}ImageNet, respectively. At $T=21$, TALON-QV achieves average accuracies of $85.24\%$, $90.10\%$, $77.83\%$, and $96.47\%$, exceeding SEC-prompt by $0.78$, $1.29$, $1.41$, and $0.57$ points, respectively. The complete comparison with other methods is reported in Table~C.4 of the supplementary material.}

\textcolor{black}{From $T=11/T=9$ to $T=21$, cumulative EKT time grows by $1.99\times$, $2.52\times$, $2.33\times$, and $2.16\times$ on CUB200, CIFAR100, ImageNet-R, and \textit{mini}ImageNet, respectively, while end-to-end wall-clock time increases by $18.4\%$, $4.3\%$, $14.7\%$, and $10.8\%$. Retained training state increases to $25.63$--$26.51$ MiB because additional Teachers and their prototypes are retained. Peak GPU memory remains unchanged because the frozen Teachers are evaluated sequentially without retaining gradients. Deployment state also remains unchanged because the accumulated Teachers and training-only components are not required for inference.}

\begin{table}[t]
\centering
\scriptsize
\setlength{\tabcolsep}{3pt}
\renewcommand{\arraystretch}{1.05}
\caption{\textcolor{black}{Training-phase comparison with baselines at $T=11$ on CUB200 and ImageNet-R and at $T=9$ on CIFAR100 and \textit{mini}ImageNet. We report training-state expanded parameters, peak allocated GPU memory, and training time per epoch. }}
\label{tab:fscil_train_cost}
\begin{tabular*}{\textwidth}{@{\extracolsep{\fill}}llccc@{}}
\toprule
Dataset & Method & \shortstack{Training\\params (M)} & \shortstack{Peak GPU\\(MiB)} & \shortstack{Time/epoch\\(s)} \\
\midrule
\shortstack[l]{CUB200\\($T=11$)} & ASP & \textbf{2.00} & 5213.16 & \underline{15.94} \\
& SEC-prompt & 4.03 & \underline{4892.90} & \textbf{3.63} \\
& TALON-MLP & \underline{3.10} & \textbf{4372.37} & 23.96 \\
& TALON-QV & 6.19 & 4961.68 & 24.36 \\
\midrule
\shortstack[l]{CIFAR100\\($T=9$)} & ASP & \underline{2.00} & 9738.25 & 80.00 \\
& SEC-prompt & \textbf{1.99} & \underline{4747.56} & \textbf{18.86} \\
& TALON-MLP & 2.51 & \textbf{4371.35} & \underline{28.15} \\
& TALON-QV & 5.01 & 4960.67 & 28.34 \\
\midrule
\shortstack[l]{ImageNet-R\\($T=11$)} & ASP & \underline{4.83} & 5692.49 & 34.45 \\
& SEC-prompt & 5.64 & 6474.20 & \textbf{7.42} \\
& TALON-MLP & \textbf{3.10} & \textbf{4372.37} & 29.39 \\
& TALON-QV & 6.19 & \underline{4961.68} & \underline{24.99} \\
\midrule
\shortstack[l]{\textit{mini}ImageNet\\($T=9$)} & ASP & \underline{2.00} & 9738.00 & 90.89 \\
& SEC-prompt & \textbf{1.99} & \underline{4747.56} & \textbf{28.66} \\
& TALON-MLP & 2.51 & \textbf{4371.35} & \underline{28.84} \\
& TALON-QV & 5.01 & 4960.67 & 29.26 \\
\bottomrule
\end{tabular*}
\end{table}

\begin{table}[t]
\color{black}
\centering
\footnotesize
\setlength{\tabcolsep}{5pt}
\caption{\textcolor{black}{Changes produced by extending TALON-QV from $T=11/T=9$ to $T=21$. EKT and retained-state columns report multiplicative growth.}}
\label{tab:talon_qv_long_sequence_change_main}
\begin{tabular*}{\textwidth}{@{\extracolsep{\fill}}lcccccc@{}}
\toprule
Dataset & Tasks & $\Delta\bar{\mathcal A}$ & \shortstack{Wall\\increase} & \shortstack{EKT\\growth} & \shortstack{Retained\\growth} & \shortstack{Gain over\\SEC-prompt} \\
\midrule
CUB200 & $11\rightarrow21$ & $+0.29$ & $+18.4\%$ & $1.99\times$ & $1.74\times$ & $+0.78$ \\
CIFAR100 & $9\rightarrow21$ & $+0.07$ & $+4.3\%$ & $2.52\times$ & $2.11\times$ & $+1.29$ \\
ImageNet-R & $11\rightarrow21$ & $-0.16$ & $+14.7\%$ & $2.33\times$ & $1.74\times$ & $+1.41$ \\
\textit{mini}ImageNet & $9\rightarrow21$ & $+0.03$ & $+10.8\%$ & $2.16\times$ & $2.11\times$ & $+0.57$ \\
\bottomrule
\end{tabular*}
\end{table}

\textcolor{black}{As shown in Table~\ref{tab:fscil_train_cost}, TALON-MLP maintains the lowest peak GPU memory usage ($\sim4.3$\,GB) across all datasets, corresponding to reductions of $7.9\%$--$32.5\%$ relative to SEC-prompt and $16.1\%$--$55.1\%$ relative to ASP. The reported training time per epoch, approximately $24$--$29$\,s, is higher than SEC-prompt because EKT performs sequential forward passes through the frozen Teacher ensemble, but remains lower than ASP on three of the four benchmarks. Moreover, EKT requires only 5 epochs, compared with 10--20 epochs for Teacher training, which limits its absolute cost at the corresponding $T=11/T=9$ task horizons.}

{\color{black}Table~\ref{tab:talon_qv_long_sequence_change_main} shows that extending TALON-QV to $T=21$ changes average accuracy by at most $0.29$ percentage points. Cumulative EKT time grows by $1.99\times$--$2.52\times$, while end-to-end wall-clock time increases by $4.3\%$--$18.4\%$ and retained training state grows by $1.74\times$--$2.11\times$. TALON-QV continues to outperform SEC-prompt by $0.57$--$1.41$ points at $T=21$. Peak GPU memory and deployment storage remain unchanged in every comparison.}

{\color{black}\textbf{Semantic-Score Computation:} In addition to the repeated Teacher forwards used by EKT, semantic coefficient construction is performed once per incremental stage using cached current-task Teacher features. The complete computation includes feature extraction, prototype construction, and similarity evaluation but requires neither learnable parameters nor backward propagation. On an NVIDIA A800, its cumulative time over a complete incremental sequence is $27.06$ s on CUB200, $11.36$ s on CIFAR100, $29.81$ s on ImageNet-R, and $14.33$ s on \textit{mini}ImageNet. These values correspond to $4.08\%$, $0.33\%$, $1.84\%$, and $0.31\%$ of task-specific Teacher-training time, respectively. Thus, semantic weighting adds a limited training-only cost and does not change TALON's trainable parameter count, deployment state, or inference complexity.}

Detailed per-method expanded parameter calculations and per-session EKT timing breakdowns are provided in the supplementary material.

\subsection{Ablation Studies}
This subsection presents comprehensive ablation studies to validate the effectiveness of individual components in TALON. We systematically investigate the contribution of each module and analyze the impact of architectural choices on overall performance.

\textbf{Different Components:}
We first evaluate the contribution of TALON’s core components, with results reported in Table~\ref{tab: ablation_studies_components}.

\begin{table*}[t]
    \centering
     \caption{Ablation studies of different components. For each metric, the left/right values represent performance with MLP-Teacher and QV-Teacher fine-tuning, respectively
        }
    \label{tab: ablation_studies_components}
    \resizebox{1.0\linewidth}{!}{
        \begin{tabular}{@{}lcccccc@{}}
            \toprule
            \multirow{2}{*}{Ablated Components} 
            & \multicolumn{2}{c}{CUB200 ($T$=11)} 
            & \multicolumn{2}{c}{CIFAR100 ($T$=9)} 
            & \multicolumn{2}{c}{\textit{mini}ImageNet ($T$=9)}\\
            \cmidrule(lr){2-3} \cmidrule(lr){4-5} \cmidrule(lr){6-7}
            & $\mathcal{A}_L$ & $\bar{\mathcal{A}}$
            & $\mathcal{A}_L$ & $\bar{\mathcal{A}}$
            & $\mathcal{A}_L$ & $\bar{\mathcal{A}}$\\
            \midrule
            w/o LoRA-Teacher KD 
            &29.13\ /\ 71.76 &45.61\ /\ 77.50
            &74.49\ /\ 85.27 &79.16\ /\ 88.08 
            &87.70\ /\ 93.60 &93.71\ /\ 95.20
            \\
            w/o Semantic Similarity
            &83.14\ /\ 83.56 &84.07\ /\ 84.02
            &82.12\ /\ 86.37 &88.52\ /\ 89.19 
            &94.12\ /\ 94.43 &95.18\ /\ 95.42
            \\
            \midrule
            TALON-MLP\ /\ QV 
            &\bf{85.24\ /\ 84.31} &\bf{85.55\ /\ 84.95}
            &\bf{87.95\ /\ 87.66} &\bf{90.26\ /\ 90.03} 
            &\bf{95.22\ /\ 95.49} &\bf{96.22\ /\ 96.44}
            \\
            \bottomrule
        \end{tabular} 
        }
\end{table*}
\begin{itemize}
    \item \textbf{w/o LoRA-Teacher KD:} This variant removes the ensemble knowledge distillation mechanism and instead relies on a single LoRA-Student continuously updated across incremental tasks. This design yields inferior performance, as it fails to balance the stability–plasticity trade-off. In contrast, assigning a dedicated LoRA-Teacher per task preserves past knowledge more effectively while supporting adaptation to new tasks, confirming the benefit of task-specific teachers in PEFT-based continual learning.
    \item \textbf{w/o Semantic Similarity:} This variant eliminates the adaptive distillation coefficients and treats all teachers equally during knowledge transfer. As shown in Table~\ref{tab: ablation_studies_components}, this consistently degrades performance across datasets. For instance, on CIFAR100, the average accuracy ($\bar{\mathcal{A}}$) of TALON-MLP drops from 90.26\% to 88.52\% (-1.74\%). Without semantic guidance, the student is unable to prioritize knowledge from relevant teachers, making it more vulnerable to conflicting supervision signals.
\end{itemize}

Together, these results highlight that both the ensemble distillation mechanism and the semantic-aware coefficients are indispensable for TALON’s superior performance in Few-Shot Class-Incremental Learning.

{\color{black}\textbf{Comparison of Teacher-Weighting Strategies:} To isolate the effect of the coefficient formulation, we keep the TALON-QV architecture, frozen Teachers, pre-EKT Student checkpoints, auxiliary head, prototype banks, data splits, class orders, optimization settings, and EKT objective fixed, and change only the computation of the Teacher coefficients. We compare summed cosine with T0-KD, Uniform-KD, maximum similarity, mean pairwise cosine, centroid cosine, normalized negative Euclidean similarity, and learnable task-level weighting.}

\begin{table*}[t]
\color{black}
\centering
\scriptsize
\setlength{\tabcolsep}{3.2pt}
\caption{\textcolor{black}{Comparison of TALON's summed semantic-guided coefficient with alternative task-level Teacher-weighting strategies. Values are average accuracy $\bar{\mathcal A}$ (\%). ``Macro Avg.'' is the unweighted mean over the four datasets, and $\Delta$ denotes the change relative to summed cosine. Macro averages and differences are calculated before rounding.}}
\label{tab:semantic_weighting_main}
\resizebox{\textwidth}{!}{\begin{tabular}{lcccccc}
\toprule
Weighting strategy & CUB200 & CIFAR100 & ImageNet-R & \textit{mini}ImageNet & Macro Avg. & $\Delta$ \\
\midrule
Summed cosine (TALON-QV) & \textbf{84.95} & \textbf{90.03} & \textbf{77.99} & 96.44 & \textbf{87.35} & -- \\
T0-KD & 84.23 & 89.11 & 76.82 & 96.44 & 86.65 & $-0.70$ \\
Uniform-KD & 84.16 & 89.27 & 76.16 & 96.46 & 86.51 & $-0.84$ \\
Maximum similarity & 84.81 & 89.09 & 76.89 & \textbf{96.47} & 86.82 & $-0.54$ \\
Mean pairwise cosine & 84.53 & 89.08 & 75.80 & 96.43 & 86.46 & $-0.89$ \\
Centroid cosine & 84.74 & 89.09 & 75.77 & 96.43 & 86.51 & $-0.85$ \\
Negative Euclidean & 83.68 & 89.09 & 75.81 & 96.43 & 86.25 & $-1.10$ \\
Learnable task-level weighting & 82.59 & 90.02 & 76.21 & 96.36 & 86.30 & $-1.06$ \\
\bottomrule
\end{tabular}}
\end{table*}

\textcolor{black}{Table~\ref{tab:semantic_weighting_main} shows that summed pairwise cosine achieves the highest four-dataset macro-average accuracy of $87.35\%$. All alternative weighting rules reduce the macro average by $0.54$--$1.10$ percentage points. TALON-QV ranks first on CUB200, CIFAR100, and ImageNet-R, while remaining within $0.03$ points of the best result on \textit{mini}ImageNet. The class-count-normalized mean score obtains $86.46\%$, corresponding to a decrease of $0.89$ points, and T0-KD obtains $86.65\%$, corresponding to a decrease of $0.70$ points. These results support summed cosine as the most effective and balanced weighting formulation across the four benchmarks.}

\textcolor{black}{The learnable task-level weighting control obtains a four-dataset macro-average accuracy of $86.30\%$, which is $1.06$ percentage points below summed cosine. Its average accuracies are lower by $2.36$ and $1.78$ points on CUB200 and ImageNet-R, respectively, while the results remain close on CIFAR100 and \textit{mini}ImageNet, with differences of $0.01$ and $0.08$ points. Under the same checkpoints and EKT settings, the evaluated learnable parameterization does not improve average accuracy on any of the four benchmarks. These results support retaining summed cosine as the default weighting rule without additional trainable Teacher-weight parameters. }

\textbf{Knowledge Consolidation Strategies:}
A central design question in TALON is \emph{how} to consolidate knowledge from multiple task-specific LoRA-Teachers. We compare TALON's EKT (output-level knowledge distillation) against three representative alternatives, directly addressing whether (i) using the teacher ensemble at inference can replace the distilled student, and (ii) simple deterministic merge baselines suffice. Results are summarized in Table~\ref{tab:fscil_router_acc}.

\begin{itemize}
    \item \textbf{Ensemble-Logits (Teacher Ensemble at Inference):} This variant retains all LoRA-Teachers and ensembles their logit outputs at test time without distilling a student. It degrades sharply on later incremental tasks: $\mathcal{A}_L$ drops to 64.90\% on CIFAR100 and 50.43\% on ImageNet-R, trailing TALON by 12.45\% and 13.15\% in $\bar{\mathcal{A}}$. Without knowledge consolidation, conflicting teacher signals accumulate, and inference cost scales linearly with the number of tasks.
    \item \textbf{Ensemble-Weights (LoRA Weight Averaging):} This baseline averages the LoRA weight matrices across all teachers: $\Delta \mathbf{W}_{\text{merged}} = \frac{1}{t+1} \sum_{k=0}^{t} \Delta \mathbf{W}_k$. On CUB200 and ImageNet-R, this produces catastrophic results ($\mathcal{A}_L$ of 6.91\% and 13.53\%), because linear parameter averaging destroys the non-linear representations learned by individual teachers. On CIFAR100 and \textit{mini}ImageNet, the damage is less severe, but $\bar{\mathcal{A}}$ still lags behind TALON by 2.32\% and 0.41\%, respectively.
    \item \textbf{Learnable Routers (MoE-style):} A layer-wise learnable router computes softmax gating weights over all LoRA-Teachers. For each Transformer layer~$l$, the router produces per-expert weights via a two-layer MLP: $\mathbf{g}^{(l)} = \text{Softmax}\!\left( \mathbf{W}_2^{(l)} \cdot \text{ReLU}(\mathbf{W}_1^{(l)} \mathbf{h}^{(l)}) \right)$, and the expert output is $\mathbf{h}'^{(l)} = \mathbf{h}^{(l)} + \sum_{k} g_k^{(l)} \cdot \mathbf{T}_k^{(l)}(\mathbf{h}^{(l)})$. Despite being standard practice in MoE literature, this approach underperforms TALON by 3.93--17.96\% in $\bar{\mathcal{A}}$, because the limited 5 samples per class are insufficient to train reliable gating networks without severe overfitting.
\end{itemize}

\begin{table*}[t]
    \centering
    \caption{Comparison of different knowledge consolidation strategies on \textbf{FSCIL}. We report the base, last, and average accuracy. All methods are based on the same pre-trained model (\textbf{ViT-B/16-IN1K}).}
    \label{tab:fscil_router_acc}
    \setlength{\tabcolsep}{1pt}
    \resizebox{\linewidth}{!}{%
    \begin{tabular}{@{}lcccccccccccc@{}}
            \toprule
            \multirow{2}{*}{Method} & 
            \multicolumn{3}{c}{CUB200 ($T$=11)} & 
            \multicolumn{3}{c}{CIFAR100 ($T$=9)} &
            \multicolumn{3}{c}{ImageNet-R ($T$=11)} &
            \multicolumn{3}{c}{\textit{mini}ImageNet ($T$=9)}
            \\
            \cmidrule(lr){2-4} \cmidrule(l){5-7} \cmidrule(l){8-10}\cmidrule(l){11-13}
            & $\mathcal{A}_{\text{Base}}\uparrow $ & $\mathcal{A}_L \uparrow$ & $\bar{\mathcal{A}} \uparrow$ 
            & $\mathcal{A}_{\text{Base}} \uparrow$ & $\mathcal{A}_L \uparrow$ & $\bar{\mathcal{A}} \uparrow$
            & $\mathcal{A}_{\text{Base}} \uparrow$ & $\mathcal{A}_L \uparrow$ & $\bar{\mathcal{A}} \uparrow$
            & $\mathcal{A}_{\text{Base}} \uparrow$ & $\mathcal{A}_L \uparrow$ & $\bar{\mathcal{A}} \uparrow$ \\
            \midrule
        Routers
            &88.56 &79.13 &81.62
            &\underline{93.17} &57.29 &72.30
            &84.47 &55.50 &64.22 
            &97.68 &93.44 &95.40
        \\
        Ensemble-Weights
            &88.73 &6.91 &48.48
            &91.28 &85.57 &87.94
            &82.86 &13.53 &48.77
            &97.52 &95.14 &96.03
        \\
        Ensemble-Logits
            &\textbf{90.35} &67.51 &76.10
            &\textbf{94.32} &64.90 &77.81
            &\textbf{87.23} &50.43 &64.84
            &\textbf{97.93} &85.44 &91.04
        \\
        \midrule
        TALON-MLP 
            &\underline{89.24} &\textbf{85.24} &\textbf{85.55} 
            &93.07 &\textbf{87.95} &\textbf{90.26} 
            &84.53 &\underline{72.23} &\underline{77.54}
            &97.68 &\underline{95.22} &\underline{96.22} 
        \\
        TALON-QV 
            &89.07 &\underline{84.31} &\underline{84.95} 
            &93.10 &\underline{87.66} &\underline{90.03} 
            &\underline{85.06} &\textbf{72.55} &\textbf{77.99}
            &\underline{97.85} &\textbf{95.49} &\textbf{96.44} 
        \\
        \bottomrule
        \end{tabular}}
\end{table*}

\textcolor{black}{These results support TALON's use of output-level knowledge distillation: by operating on soft logits rather than directly merging parameters, EKT consolidates the task-specific Teacher knowledge into a single compact Student for efficient inference. The task-level semantic-guided coefficients emphasize Teachers relevant to the current task at each incremental stage; the same coefficient vector is shared by all samples and EKT epochs in that stage.}

\begin{table}[t]
    \caption{Results of different methods on CIFAR100 ($T$=9) using {\bf supervised} pre-training model  ViT-B/16-IN21K (denoted as Sup-21K) and {\bf self-supervised} model DINO  %
    }
    \label{tab:vary_pre-trained}
    \centering
    \small
    \setlength{\tabcolsep}{3pt}  
    
    \begin{tabular}{c|l|ccc} 
        \hline
        \rule{0pt}{10pt} PTM & Method & $\mathcal{A}_{\text{Base}}$ & $\mathcal{A}_L$ & $\bar{\mathcal{A}}$  \\
        \hline
        \multirow{11}{*}{ Sup-21K} & 
        Full Finetune 
        &90.97  &40.83  &63.89\\
        & SimpleFSCIL 
        &83.00  &71.53 &76.71 \\
        & L2P 
        &91.92  &55.35 &71.09  \\
        & CODA-Prompt
        &93.23  &55.89  &71.87 \\
        & InfLoRA
        &\bf{94.10}  &56.18  &72.29 \\
        & SD-LoRA
        &93.58  &78.56  &84.98 \\
        & ASP
        &92.33  &\bf{87.83}  &89.73 \\
        & SEC-prompt
        &93.52  &84.57  &89.02 \\
        & TALON-MLP 
        &93.52  &87.62  &90.10 \\
        & TALON-QV 
        &93.43  &87.42  &\bf{90.24} \\
        \hline
        \multirow{11}{*}{DINO} & 
        Full Finetune 
        &88.27  &41.36  &55.61\\
        & SimpleFSCIL 
        &58.82  &42.49  &49.46\\
        & L2P 
        &85.65  &48.68  &64.23\\
        & CODA-Prompt
        &\bf{89.50}  &53.73  &68.86\\
        & InfLoRA
        &88.52 &52.87  &68.11\\
        & SD-LoRA
        &88.57  &30.12  &27.75\\
        & ASP
        &68.23  &48.69  &56.98\\
        & SEC-prompt
        &85.40  &69.47  &76.99\\
        & TALON-MLP 
        &87.13  &74.85  &80.12\\
        & TALON-QV 
        &87.32  &\bf{75.87} &\bf{80.76} \\
        \hline
    \end{tabular}
\end{table}

\textbf{Robustness across Pre-training Paradigms:} Table~\ref{tab:vary_pre-trained} provides a comprehensive comparison across supervised (ViT-B/16-IN21K) and self-supervised (DINO) pre-training. Key findings:
\begin{itemize}
    \item \textbf{Supervised (IN21K):} TALON-QV achieves $\bar{\mathcal{A}}$ = 90.24\%, outperforming ASP (89.73\%) and SEC-prompt (89.02\%)
    \item \textbf{Self-supervised (DINO):} TALON-QV achieves $\bar{\mathcal{A}}$ = 80.76\%, significantly outperforming SEC-prompt (76.99\%) and ASP (56.98\%)
\end{itemize}
Notably, prompt-based methods (ASP, SEC-prompt) are more sensitive to pre-training paradigm shifts, while TALON maintains robust performance across different feature spaces.

Different PTMs and different LoRA-Teachers are further analyzed in the supplementary material.

\subsection{Visualization of Incremental Sessions} 

\begin{figure}
    \centering
    \subfloat[First stage\label{fig:tsne1}]{\includegraphics[width=0.48\linewidth]{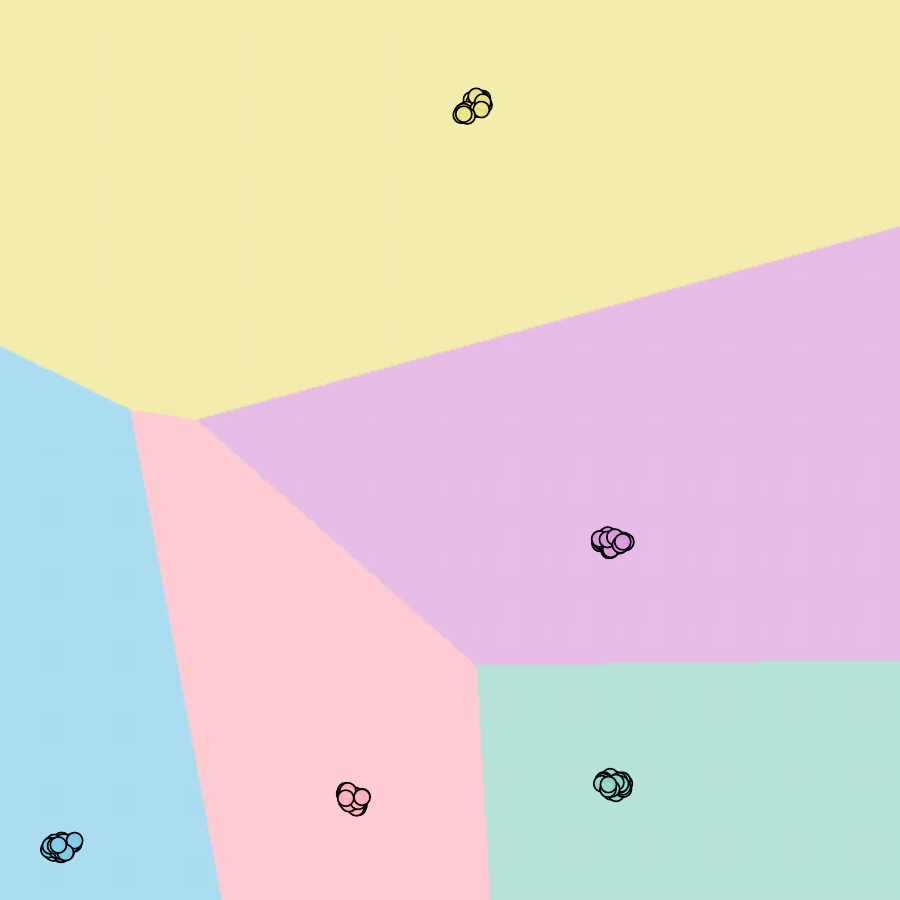}}%
    \hfil
    \subfloat[Second stage\label{fig:tsne2}]{\includegraphics[width=0.48\linewidth]{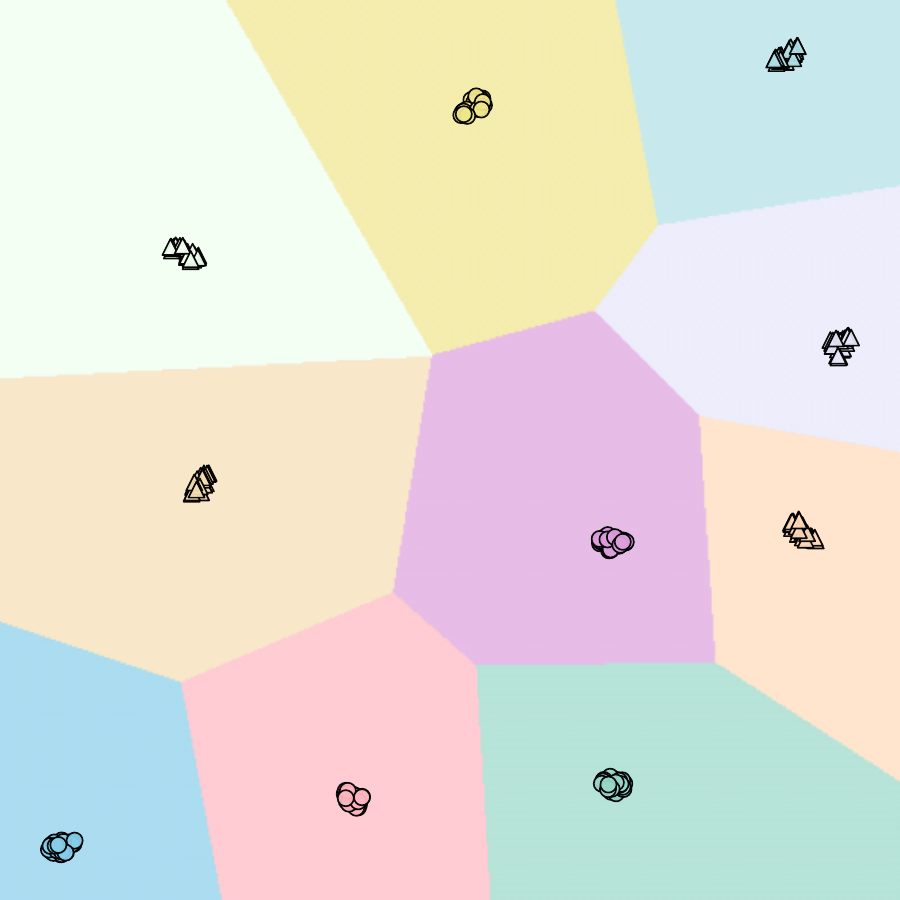}}
    \caption{Visualization of the decision boundary on \textit{mini}ImageNet between two incremental tasks. Dots represent old classes, and triangles stand for new classes. Decision boundaries are shown with the shadow region.}
    \label{fig:tsne}
\end{figure}

To validate TALON's effectiveness in the few-shot regime, we visualize feature representations from the first and second incremental stages (following the base stage) on \textit{mini}ImageNet using t-SNE~\cite{van2008visualizing}, as shown in Fig.~\ref{fig:tsne}. Two key observations emerge: (i) TALON effectively separates instances into their corresponding classes with well-defined decision boundaries, and (ii) when extending from the first to the second stage, TALON maintains clear separation for both old classes (dots) and new classes (triangles), demonstrating effective knowledge consolidation without catastrophic forgetting. These visualizations confirm that TALON learns genuinely discriminative features rather than relying solely on increased model capacity.


\begin{figure}[t]
    \begin{center}
        \includegraphics[width=1\linewidth]{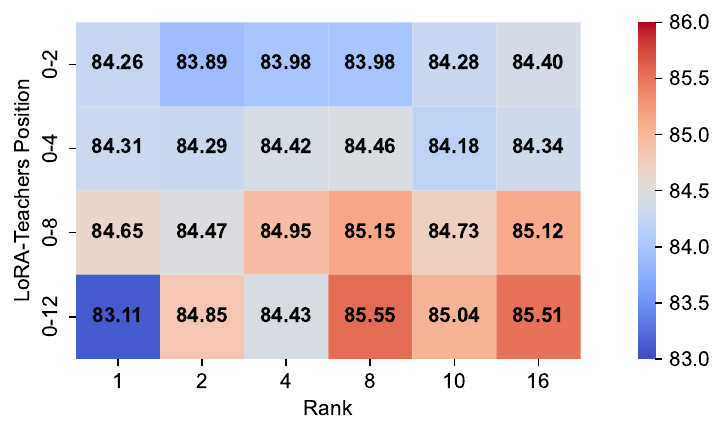}
    \end{center}
    \caption{Sensitivity of hyperparameters (the rank $r$, the insertion positions) on CUB200 ($T$=11).}
    \label{fig:parameter-analysis-rank}
\end{figure}

\subsection{Further Analysis}

\textbf{Parameter Sensitivity Analysis:}
TALON involves two key hyperparameters: (1) the rank $r$ of the LoRA-Teachers, and (2) their insertion positions within the Vision Transformer (ViT).

To examine the sensitivity of these hyperparameters, we conducted experiments on the CUB200 ($T$=11) benchmark. Specifically, the rank $r$ was varied over $\{1, 2, 4, 8, 10, 16\}$, while the insertion positions were configured as \{0-2, 0-4, 0-8, 0-12\}, where ``0-2'' indicates insertion into the first two Transformer layers. The results, illustrated in Fig.~\ref{fig:parameter-analysis-rank}, show that TALON delivers stable performance across a wide range of settings, indicating robustness to variations in both rank and insertion position. Although reported here on CUB200, similar trends are observed on other datasets. Based on this analysis, we adopt $r=8$ and insert LoRA-Teachers into all Transformer layers (0-12) as the default configuration. 

{\color{black}The ``Different KD Losses'' section of the supplementary material compares Logit-KL, Logit-L2, Feature-L2, Feature-CS, and feature-space-consistent Prototype Relation for both TALON-MLP and TALON-QV. Logit-KL achieves the highest combined four-dataset macro-average of 87.37\%. Although several alternatives are close in individual settings, none provides a consistent improvement across datasets and variants. These results support Logit-KL as TALON's most balanced EKT objective. }

\begin{table}[t]
\color{black}
\centering
\footnotesize
\setlength{\tabcolsep}{1pt}
\caption{\textcolor{black}{Prototype-drift analysis and comparison of historical-prototype update strategies. $D_{\mathrm{drift}}$ is the final-session mean cumulative L2 drift; $\rho_{\Delta}$ is the pooled drift-direction agreement; and $R_{\mathrm{cent}}$ is the pooled centroid-distance reduction. $\Delta_{\mathrm{SDC}}$ and $\Delta_{\mathrm{Recomp}}$ are mean paired changes in final accuracy relative to Frozen, in percentage points. $D_{\mathrm{drift}}$ and Frozen $\mathcal{A}_L$ are reported as mean$\pm$standard deviation over class-order seeds 42, 1993, and 2025.}}
\label{tab:prototype_drift_sdc}
\begin{tabular*}{\textwidth}{@{\extracolsep{\fill}}llcccccc@{}}
\toprule
Dataset & Variant & $D_{\mathrm{drift}}$ & $\rho_{\Delta}$ & $R_{\mathrm{cent}}$ & \shortstack{Frozen\\$\mathcal{A}_L$ (\%)} & $\Delta_{\mathrm{SDC}}$ & $\Delta_{\mathrm{Recomp}}$ \\
\midrule
CUB200 & TALON-MLP & $0.1105\pm0.0188$ & 0.860 & 24.0\% & $83.90\pm1.08$ & $+0.042$ & $+0.014$ \\
& TALON-QV & $0.0605\pm0.0237$ & 0.810 & 18.7\% & $83.80\pm0.85$ & $-0.042$ & $-0.028$ \\
\midrule
CIFAR100 & TALON-MLP & $0.0503\pm0.0057$ & 0.852 & 25.9\% & $87.42\pm0.66$ & $-0.003$ & $+0.003$ \\
& TALON-QV & $0.0130\pm0.0032$ & 0.650 & 5.9\% & $87.48\pm0.23$ & $+0.030$ & $+0.020$ \\
\midrule
ImageNet-R & TALON-MLP & $0.1305\pm0.0279$ & 0.874 & 25.7\% & $73.59\pm0.80$ & $+0.078$ & $-0.044$ \\
& TALON-QV & $0.0666\pm0.0040$ & 0.766 & 11.4\% & $73.66\pm1.21$ & $-0.017$ & $-0.022$ \\
\midrule
\textit{mini}ImageNet & TALON-MLP & $0.0143\pm0.0084$ & 0.841 & 24.6\% & $95.30\pm0.23$ & $-0.010$ & $+0.013$ \\
& TALON-QV & $0.00215\pm0.00060$ & 0.627 & 3.5\% & $95.44\pm0.36$ & $+0.007$ & $+0.010$ \\
\bottomrule
\end{tabular*}
\end{table}

\subsection{\textcolor{black}{Prototype--Feature Consistency Across Incremental Stages}}
\label{sec:prototype_drift}

{\color{black}
For a class $c$ introduced at session $j$, let $\mathbf{P}_\text{Stu}^{(j)}[c]$ denote its stored prototype and $\widetilde{\mathbf{P}}_\text{Stu}^{(t)}[c]$ the centroid recomputed from the same historical samples using the current Student $\mathbf{S}_t$. Historical samples are accessed only for post-hoc drift measurement and prototype recomputing; they are never used by TALON or SDC.

Following SDC~\cite{yu2020semantic}, we estimate historical-class displacement using current-task features before and after the Student update:
\begin{equation}
\mathbf{z}_{i,t}^{-}=\operatorname{norm}\!\left(\phi(\mathbf{x}_i;\mathbf{S}_{t-1})\right),\qquad \mathbf{z}_{i,t}^{+}=\operatorname{norm}\!\left(\phi(\mathbf{x}_i;\mathbf{S}_t)\right).
\label{eq:sdc_features}
\end{equation}
The historical prototype is recursively updated as
\begin{equation}
\begin{aligned}
w_{i,c}^{(t)}&=\exp\!\left(-\frac{\left\|\mathbf{z}_{i,t}^{-}-\mathbf{P}_{\mathrm{SDC}}^{(t-1)}[c]\right\|_2^2}{2\sigma^2}\right),\\
\mathbf{P}_{\mathrm{SDC}}^{(t)}[c]&=\mathbf{P}_{\mathrm{SDC}}^{(t-1)}[c]+\frac{\sum_{(\mathbf{x}_i,y_i)\in\mathcal{D}_t}w_{i,c}^{(t)}\left(\mathbf{z}_{i,t}^{+}-\mathbf{z}_{i,t}^{-}\right)}{\sum_{(\mathbf{x}_i,y_i)\in\mathcal{D}_t}w_{i,c}^{(t)}}.
\end{aligned}
\label{eq:sdc_update}
\end{equation}
Here, $\sigma=0.20$ and $\mathbf{P}_{\mathrm{SDC}}^{(j)}[c]=\mathbf{P}_\text{Stu}^{(j)}[c]$.

The true and SDC-estimated cumulative drift vectors are
\begin{equation}
\boldsymbol{\Delta}_{c}^{j\rightarrow t}=\widetilde{\mathbf{P}}_\text{Stu}^{(t)}[c]-\mathbf{P}_\text{Stu}^{(j)}[c],\qquad \widehat{\boldsymbol{\Delta}}_{c}^{j\rightarrow t}=\mathbf{P}_{\mathrm{SDC}}^{(t)}[c]-\mathbf{P}_\text{Stu}^{(j)}[c].
\label{eq:prototype_drift_vectors}
\end{equation}
We report
\begin{equation}
\begin{aligned}
D_{\mathrm{drift}}&=\frac{1}{|\Omega_L|}\sum_{(s,c,j,L)\in\Omega_L}\left\|\boldsymbol{\Delta}_{s,c}^{j\rightarrow L}\right\|_2,\\
\rho_{\Delta}&=\frac{1}{|\Omega|}\sum_{(s,c,j,t)\in\Omega}\frac{\left\langle\widehat{\boldsymbol{\Delta}}_{s,c}^{j\rightarrow t},\boldsymbol{\Delta}_{s,c}^{j\rightarrow t}\right\rangle}{\left\|\widehat{\boldsymbol{\Delta}}_{s,c}^{j\rightarrow t}\right\|_2\left\|\boldsymbol{\Delta}_{s,c}^{j\rightarrow t}\right\|_2+\varepsilon}.
\end{aligned}
\label{eq:prototype_drift_metrics}
\end{equation}
Here, $s$ indexes the class-order seed, $\Omega$ contains all valid historical-class/session/seed records, and $\Omega_L$ denotes its final-session subset.

The relative reduction in prototype-to-current-centroid distance is
\begin{equation}
R_{\mathrm{cent}}=1-\frac{\sum_{(s,c,j,t)\in\Omega}\left\|\mathbf{P}_{\mathrm{SDC},s}^{(t)}[c]-\widetilde{\mathbf{P}}_{\text{Stu},s}^{(t)}[c]\right\|_2}{\sum_{(s,c,j,t)\in\Omega}\left\|\mathbf{P}_{\text{Stu},s}^{(j)}[c]-\widetilde{\mathbf{P}}_{\text{Stu},s}^{(t)}[c]\right\|_2}.
\label{eq:centroid_distance_reduction}
\end{equation}

Within each run, we fix the Student checkpoint, test features, class order, and cosine nearest-prototype classifier, and change only the historical prototype entries. We compare Frozen, SDC, and Historical-Data Prototype Recomputing. The last strategy recomputes each historical prototype from its original class samples using the current Student and is included only as a post-hoc diagnostic.
}

{\color{black}
Table~\ref{tab:prototype_drift_sdc} shows that SDC consistently tracks and reduces the measured displacement: $\rho_{\Delta}$ ranges from $0.627$ to $0.874$, and $R_{\mathrm{cent}}$ ranges from $3.5\%$ to $25.9\%$. Crucially, TALON's predictions remain stable despite this geometric drift. SDC changes final accuracy by only $-0.042$ to $+0.078$ percentage points, while Historical-Data Prototype Recomputing changes it by $-0.044$ to $+0.020$ points. Across all 24 runs, neither strategy produces a significant final-session McNemar result.

These findings distinguish measurable feature-space displacement from decision-level degradation. Although historical prototypes drift relative to the current Student space, TALON's cosine nearest-prototype classifier remains robust. We therefore retain Frozen, which preserves the strict exemplar-free protocol without sacrificing meaningful classification performance.
}

\subsection{\textcolor{black}{Historical-Knowledge Preservation During EKT}}
\label{sec:ekt_preservation}

{\color{black}To directly quantify historical-knowledge preservation during EKT, we record historical-task accuracy immediately before and after each EKT update. At stage $t$, we define}
\begin{equation}
{\color{black}\Delta_{\mathrm{old}}^{(t)}=\frac{1}{t}\sum_{j=0}^{t-1}\left(a_{t,j}^{\mathrm{after}}-a_{t,j}^{\mathrm{before}}\right),}
\label{eq:ekt_old_delta}
\end{equation}
{\color{black}The mean historical-task changes after Logit-KL EKT are $+0.05$, $-0.00$, and $-0.12$ percentage points on CUB200, CIFAR100, and ImageNet-R, respectively, showing that historical-task performance remains essentially unchanged on average. Comparisons with alternative EKT targets are provided in the ``Different KD Losses'' section of the supplementary material.}

\section{Conclusion}
In this work, we introduced TALON, a novel framework addressing the core challenges of Few-Shot Class-Incremental Learning (FSCIL) through task-adaptive LoRA-Teacher modules and ensemble knowledge transfer. TALON effectively balances the stability-plasticity trade-off by dynamically allocating a dedicated LoRA-Teacher for each incremental task while consolidating their collective knowledge into a unified LoRA-Student model for efficient inference.

The key contributions of TALON are:
\begin{itemize}
    \item Dynamic multi-expert paradigm: Eliminates the need for predefined module counts and achieves superior plasticity through complete parameter isolation;

    \item Distills knowledge from multiple teachers into a single student model, removing computational overhead during inference;

    \item Semantic-guided distillation: Enables optimal integration of teacher knowledge while mitigating catastrophic forgetting and overfitting.
\end{itemize}

\textcolor{black}{Extensive experiments over three class-order runs show that TALON achieves \textcolor{black}{comparable or better} mean average accuracy across four benchmark datasets. TALON maintains clear mean improvements on CIFAR100, ImageNet-R, and \textit{mini}ImageNet, while obtaining performance comparable to SEC-prompt on the more class-order-sensitive CUB200 benchmark. In addition, TALON retains its parameter-efficient design and fast single-model inference. The expanded ablation studies further show limited sensitivity to the semantic temperature, soft-distillation weighting, and EKT duration.}

Looking ahead, TALON’s dynamic adaptation and efficient knowledge consolidation provide promising directions for future incremental learning research, especially in scenarios demanding both sample efficiency and computational efficiency.

\section*{Acknowledgments}
This work is supported in part by the National Natural Science Foundation of China (Grant No. 62372054, 62006005) and National Key Research and Development Program of China (Grant No. 2022YFC3302200).


\end{document}